\pdfoutput=1

\documentclass[11pt]{article}

\usepackage{naacl2021}

\usepackage{times}
\usepackage{latexsym}

\usepackage[T1]{fontenc}

\usepackage[utf8]{inputenc}

\usepackage{microtype}

\usepackage{graphicx}
\usepackage{booktabs}
\usepackage{caption}
\usepackage{subcaption}
\usepackage{tabularx, colortbl}
\usepackage{arydshln}

\title{Attention Head Masking for Inference Time Content Selection \\in Abstractive Summarization}

\author{Shuyang Cao \and Lu Wang \\
  Computer Science and Engineering \\
  University of Michigan \\
  Ann Arbor, MI \\
  \texttt{\{caoshuy, wangluxy\}@umich.edu} \\}

\begin{document}
\maketitle

\begin{abstract}

How can we effectively inform content selection in Transformer-based abstractive summarization models? 
In this work, we present a simple-yet-effective \textit{attention head masking} technique, which is applied on encoder-decoder attentions to pinpoint salient content at inference time. 
Using attention head masking, we are able to reveal the relation between encoder-decoder attentions and content selection behaviors of summarization models. 
We then demonstrate its effectiveness on three document summarization datasets based on both \textit{in-domain} and \textit{cross-domain} settings. 
Importantly, our models outperform prior state-of-the-art models on CNN/Daily Mail and New York Times datasets. 
Moreover, our inference-time masking technique is also data-efficient, requiring less than 20\% of the training samples to outperform BART fine-tuned on the full CNN/DailyMail dataset. 

\end{abstract}
\section{Introduction}

Large pre-trained Transformers have achieved state-of-the-art results on various summarization datasets with a fine-tuning phase to streamline the summarization pipeline~\cite{lewis-etal-2020-bart,yan2020prophetnet}. 
Yet, it is still unclear \textit{how one can use large models more effectively for abstractive summarization }. 
For example, prior work shows that informing content selection via attention weight updating in recurrent neural networks can further boost summarizer performance~\cite{gehrmann-etal-2018-bottom}. However, with multi-heads attentions at all layers in Transformers~\cite{NIPS2017_3f5ee243}, highlighting salient content becomes non-trivial. 

In this work, we propose an \textbf{inference-time attention head masking mechanism} that works on encoder-decoder attentions to underscore salient content from the source and improve the quality of abstractive summaries. 
Based on this mechanism, we first demonstrate the relation between encoder-decoder attentions and content selection behaviors, on three summarization datasets of CNN/DailyMail (CNN/DM), New York Times (NYT), and XSum. 
Second, we study whether multiple heads at the same layer collectively guide the summarization. Partial masking is found to be most effective, indicating a strong collaborative effect and the importance of head selection. 

Based on these observations, we evaluate attention head masking on summarization benchmarks with salience labels provided by externally trained content selectors. 
On all three datasets, our model consistently outperforms fine-tuned BART~\cite{lewis-etal-2020-bart} and several top performing Transformer-based abstractive summarization models~\cite{zhang2019pegasus,yan2020prophetnet}.
Summaries generated by our model are also considered to have better informativeness by human judges.
Moreover, we illustrate that attention head masking is \textbf{data-efficient}: on CNN/DM, BART fine-tuned on less than $20\%$ of the training data outperforms a version trained on the full set. 
Finally, we show that our method is effective under a \textbf{cross-domain} setting. With a content selector trained on NYT, BART fine-tuned on CNN/DM gains more than three points of ROUGE scores when tested on NYT articles.\footnote{Our code is available at: \url{https://shuyangcao.github.io/projects/inference_head_masking}.}

\section{Related Work}

\noindent \textbf{Large Pre-trained Models for Summarization.} 
Many recent advancements in text summarization have been achieved by large pre-trained language models~\cite{zhang-etal-2019-pretraining,liu-lapata-2019-text,song19d,zhang2019pegasus}. 
In particular, BART has demonstrated impressive performance on summarization, and is used as the base model in this work. 
Nonetheless, all prior attempts take pre-trained models as is and conduct fine-tuning on target datasets, without knowing if it is the most effective usage.
In contrast, we bring insights into the relation between attentions and content selection via masking operations to further improve summarization performance.

\smallskip
\noindent \textbf{Content Selection for Abstractive Summarization.} 
Content selection is a crucial step, where salient information is first detected and then summarized into concise abstracts~\cite{chen-bansal-2018-fast,xu-durrett-2019-neural}. 
To minimize the propagation of selection errors,
content selection is modeled as an extra component and learned within an end-to-end trained model~\cite{Zhou_2017,li-etal-2018-improving,gehrmann-etal-2018-bottom}. 
To the best of our knowledge, we are the first to apply masks on selected layers and attention heads in Transformers for content selection in summarization. Moreover, our masking mechanism is only activated during inference, without any model modification.

\smallskip
\noindent \textbf{Analyzing Multi-head Attentions} has attracted growing interests in the NLP community~\cite{clark-etal-2019-bert,kovaleva-etal-2019-revealing}. 
Among the work that is relevant to encoder-decoder attentions, \citet{NEURIPS2019_2c601ad9} and \citet{voita-etal-2019-analyzing} observe that only a small portion of heads is relevant for translation and encoder-decoder attentions tend to be more important than self-attentions. 
Meanwhile, word alignments for machine translation are induced from encoder-decoder attention weights~\cite{li-etal-2019-word, kobayashi2020attention}. 
However, none of prior work employs attentions to improve generation quality. 
As far as we are aware, this is the first work that studies the content selection effects of encoder-decoder attentions and uses them to guide better summary generation.

\section{Attention Head Masking} 
We adopt large pre-trained sequence-to-sequence Transformer models (BART, specifically) for abstractive summarization.
Transformer is built with {\bf multi-head attentions}. Attentions are computed per step based on a query $\mathbf{q}$ along with the key and value matrices, $\mathbf{K}$ and $\mathbf{V}$:

\vspace{-2mm}
{
    \fontsize{10}{11}\selectfont
    \begin{equation}
        \mathrm{Attention} (\mathbf{q}, \mathbf{K}, \mathbf{V}) = \mathrm{softmax} (\frac{\mathbf{q} \mathbf{K}^T}{\sqrt{d_k}} + \mathbf{m}) \mathbf{V} \label{eq:attn_out} 
    \end{equation}
}%
where $d_k$ is a scaling factor and $\mathbf{m}$ is for padding or masking future tokens (when the value is $-\infty$).

\begin{figure}
    \hspace{2mm}
    \includegraphics[width=0.45\textwidth]{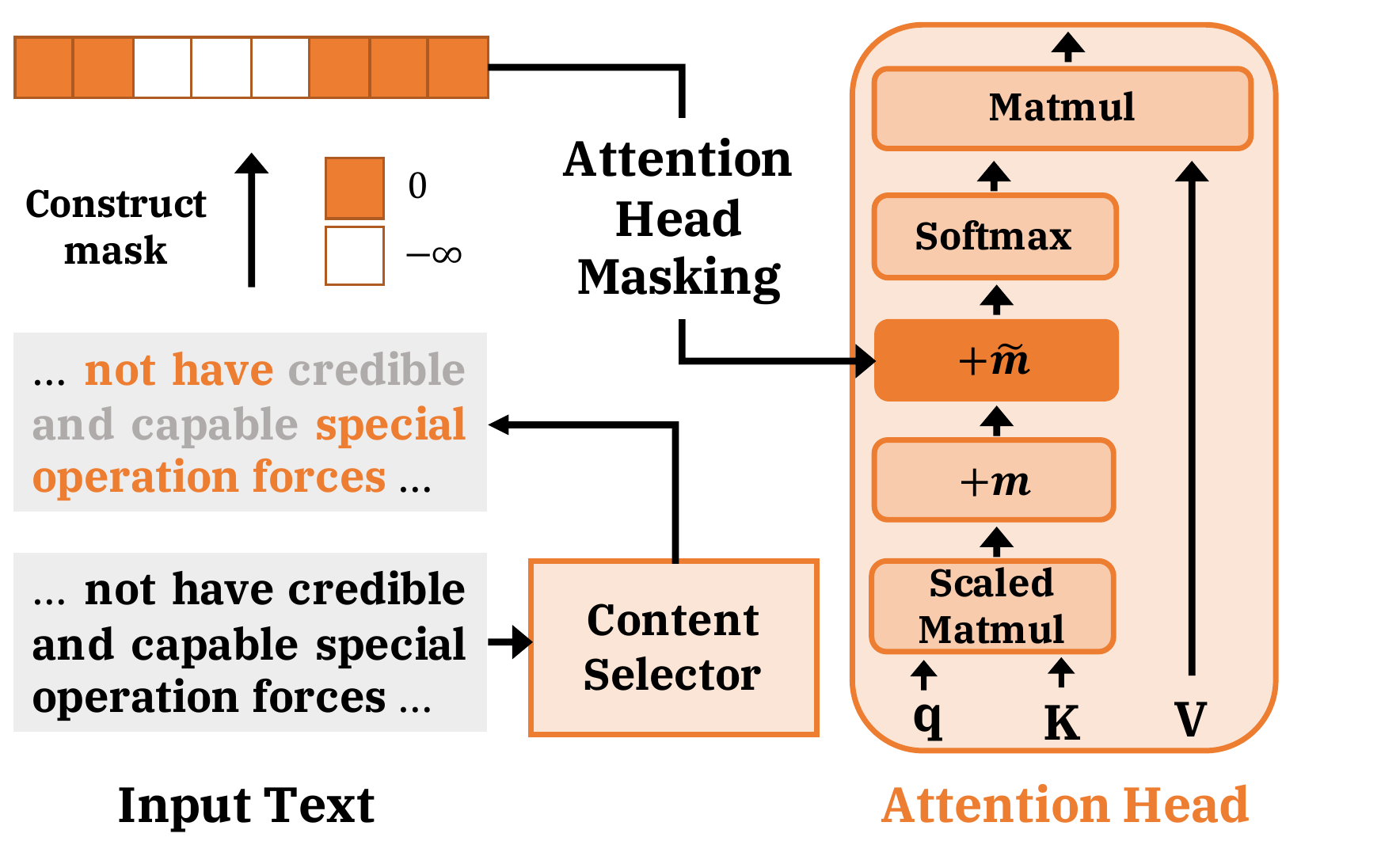}
    \caption{Illustration of attention head masking ($\mathbf{\tilde{m}}$).}
    \label{fig:model}
\end{figure}

\smallskip
\noindent \textbf{Masking Operation.} 
We propose attention head masking in encoder-decoder attentions, which blocks attentions to unimportant tokens, to better concentrate multi-head attentions on salient input tokens. Importantly, it is activated during inference.
Concretely, we add an $\mathbf{\tilde{m}}$ inside the softmax operator of Eq.~\ref{eq:attn_out}, with implementation displayed in Fig.~\ref{fig:model}. 
The size of $\mathbf{\tilde{m}}$ is the same as the input length. If the $i$-th token is tagged as salient, the corresponding element in $\mathbf{\tilde{m}}$ is set to $0$ (attendable to the attention heads), and $-\infty$ otherwise (hidden from these heads). The saliency labels can be predicted by an externally trained content selector.

\section{Encoder-decoder Attentions and Content Selection}\label{sec:analysis}

In this section, we first probe into the content selection behavior of each single head (\S~\ref{sec:content_selection}), and then study the synergism among heads at the same layer (\S~\ref{subsec:synergism}). In \S~\ref{subsec:attnfocus}, we analyze the attentions' focus. 

Our analysis is conducted on CNN/DM~\cite{NIPS2015_5945}, NYT~\cite{linguistic2008new}, and XSum~\cite{narayan-etal-2018-dont}. 
We follow \citet{lewis-etal-2020-bart} for data preprocessing and train/validation/test splits on CNN/DM and XSum, and adopt the setups in \citet{paulus2018a} for NYT, except that we keep entities and numbers.
The number of samples in training, validation, and test set are: $287{,}188$, $13{,}367$ and $11{,}490$ for CNN/DM; $588{,}909$, $32{,}716$ and $32{,}703$ for NYT; $204{,}045$, $11{,}332$ and $11{,}334$ for XSum. 

For experiments in this section, we create an \textit{analysis set} of $1{,}000$ random samples from the validation split of each dataset to reduce computational cost.

\subsection{Content Selection Effects}
\label{sec:content_selection}
First, we study the feasibility of using encoder-decoder attentions to inform content selection and subsequently boost summary informativeness. 
Concretely, we apply attention head masking based on oracle content selection labels (henceforth \textbf{oracle masking}). 
Oracle labels are constructed by aligning a reference summary to the source article, where we iteratively find the longest common subsequences between the two.

Taking a fine-tuned BART model, we apply oracle masking on each head at each layer when decoding on the analysis set. The ROUGE score obtained in this setting is denoted as $r_{ora}$. 
We then apply uniform encoder-decoder attention weights over the source to build a baseline that mimics no content selection, inspired by~\newcite{wiegreffe-pinter-2019-attention}. This yields a ROUGE score of $r_{uni}$. The \textbf{content select effect} per head can thus be calculated as the ROUGE improvement, i.e., $r_{ora} - r_{uni}$.

\begin{figure}[t]
    \centering
    \centering
    \includegraphics[width=0.44\textwidth]{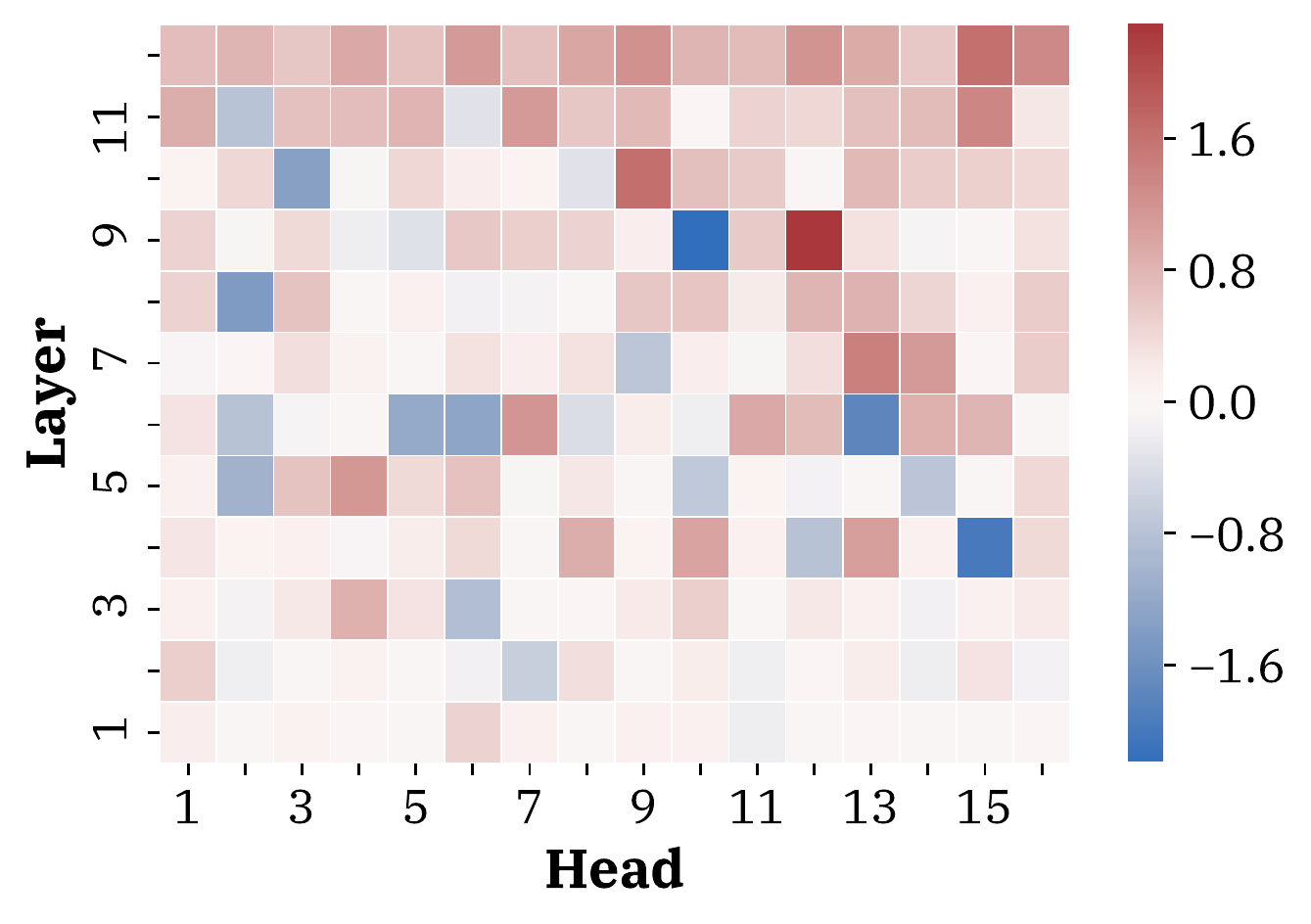}
    \caption{
    ROUGE-1 F1 improvement with oracle masks for each head at each layer on the analysis set of CNN/DM. Overall, top layers see greater improvement than bottom layers. Layer 1 is the bottom layer connected with the word embeddings. 
    }
    \label{fig:cnndm_head_perf}
\end{figure}

Overall, it is more effective to constrain attentions to salient content at the top layers, according to the results on CNN/DM in Fig.~\ref{fig:cnndm_head_perf}. Specifically, with oracle masking, the top layer yields the most ROUGE-1 improvement. We observe similar trends on NYT and XSum (figures are in Appendix~\ref{append:content_selection_effect}). This indicates the feasibility of \textbf{leveraging attention head masking to improve summary informativeness.}

\subsection{Synergism Analysis}
\label{subsec:synergism}

\begin{figure}[t]
    \centering
    \includegraphics[width=0.48\textwidth]{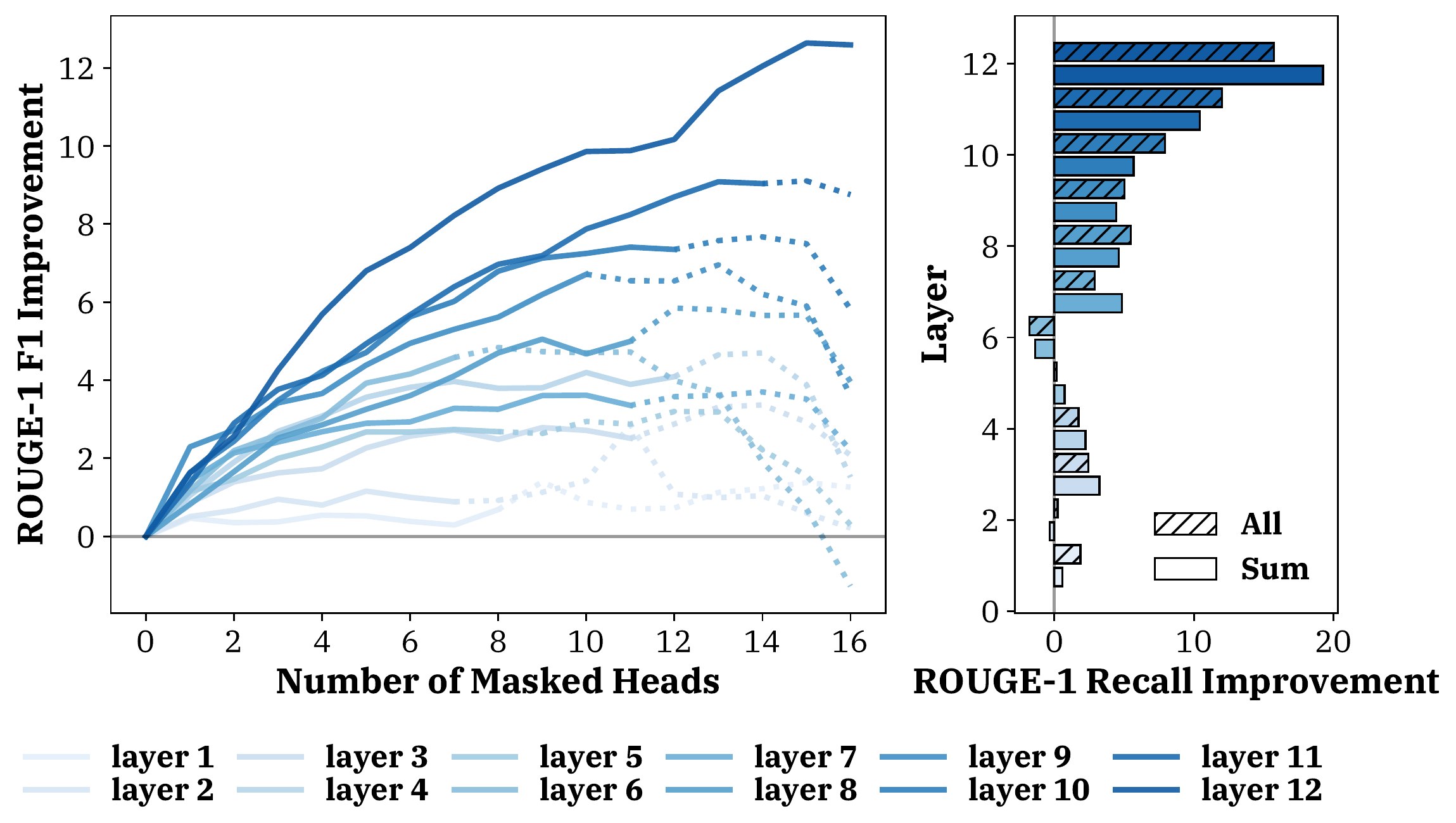}
    \vspace{-4mm}
    \caption{[\textbf{Left}] ROUGE-1 F1 improvement by incrementally applying oracle masking to the next head with most ROUGE improvement per layer on CNN/DM. Dotted lines indicate that the newly masked heads do not have individual ROUGE improvements. 
    [\textbf{Right}] ROUGE-1 recall improvement by masking all heads vs. sum of improvement by masking each head separately on CNN/DM. 
    Best viewed in color. 
    }
    \label{fig:cnndm_synergy}
\end{figure}

Next, we study whether masking multiple heads can further boost content selection and whether they form synergy. 
On the left of Fig.~\ref{fig:cnndm_synergy}, we show content selection effect by gradually applying oracle masking on more heads at each layer, with heads sorted based on individual ROUGE improvements. Notably, the most ROUGE-1 improvement is achieved by masking 15 (out of 16) heads at the top layer, suggesting \textbf{a strong collaborative effect on content selection by masking multiple heads}. 

We further compare the ROUGE score gains between oracle masking on all heads and the sum of individual effects, illustrated on the right of Fig.~\ref{fig:cnndm_synergy}. The discrepancies between the two values suggest that the heads may not be independent at pinpointing salient content. 
In Appendix~\ref{append:synergism_analysis}, we reach similar results on NYT and XSum. 

Based on the above observations, we argue that \textbf{it is necessary to select layers and heads accordingly to achieve the best content selection effect}, with more summarization results reported in \S~\ref{sec:exp_result}.

\subsection{Attention Focus} 
\label{subsec:attnfocus}

We further provide a fine-grained study on what types of words the heads attend to. 
Concretely, we consider each word generated during decoding, denoted as $y$. 
Given an attention head, we follow the highest attention weight to identify the input word $x$ (``\textbf{attendee}''). 
We study several categories of attendee $x$: 
(1) word in the reference (\textsc{salient}); 
(2) \textsc{content} word; 
(3) the \textsc{first} and \textsc{last} words in the document. 
For \textsc{salient} and \textsc{content}, we further consider two subcategories: $x = y$ (\textsc{copy}) and $x \ne y$ (\textsc{non-copy}).
We then tally the occurrences of each type of attendees per head at each layer on the analysis set. 

\begin{figure}[t]
     \vspace{-3mm}
     \begin{subfigure}[b]{0.25\textwidth}
         \includegraphics[width=\textwidth]{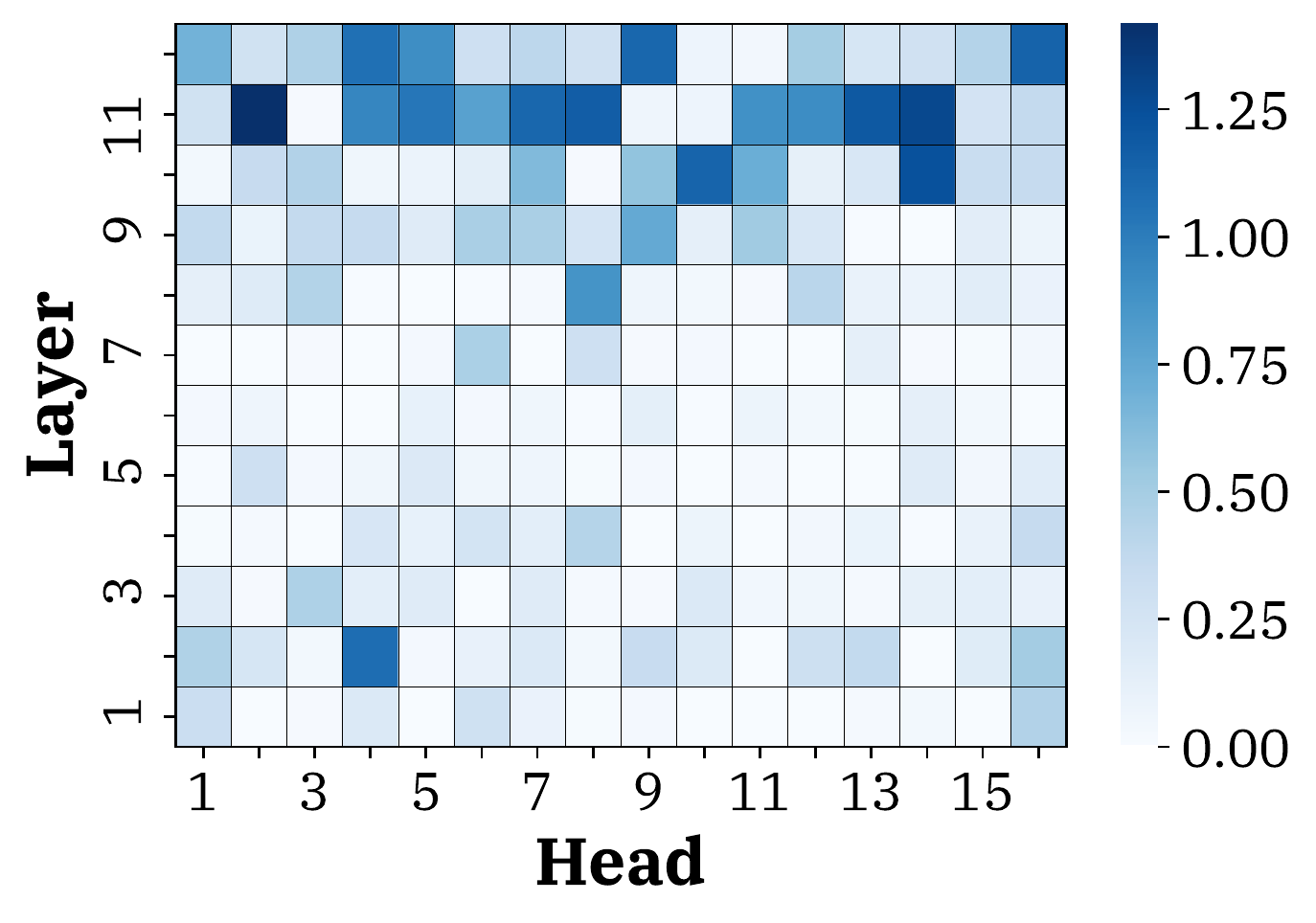}
         \caption{\textsc{copy salient}}
     \end{subfigure}\hspace*{-0.4em}
     \begin{subfigure}[b]{0.25\textwidth}
         \includegraphics[width=\textwidth]{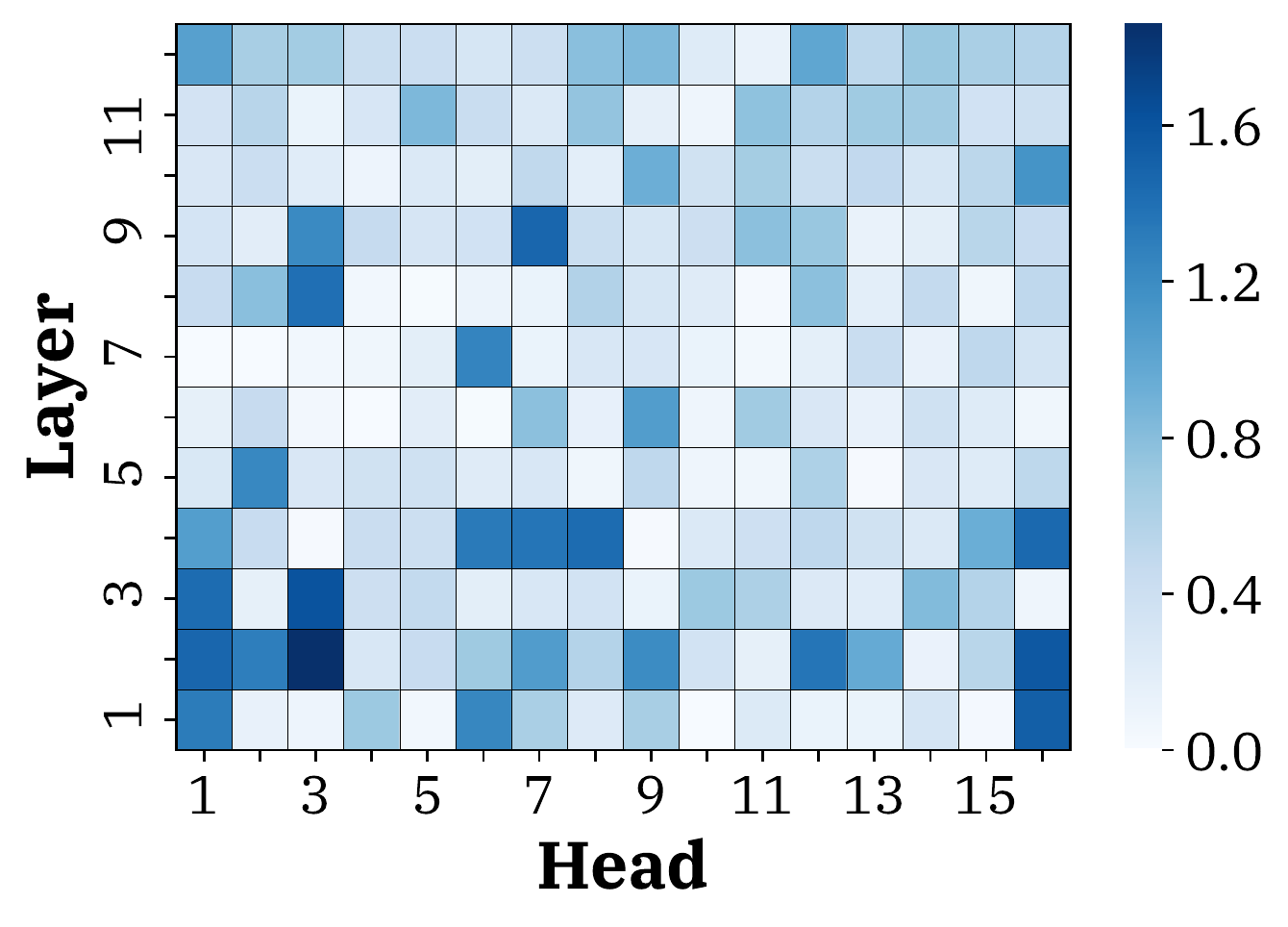}
         \caption{\textsc{non-copy salient}}
     \end{subfigure}
     \begin{subfigure}[b]{0.24\textwidth}
         \centering
         \includegraphics[width=\textwidth]{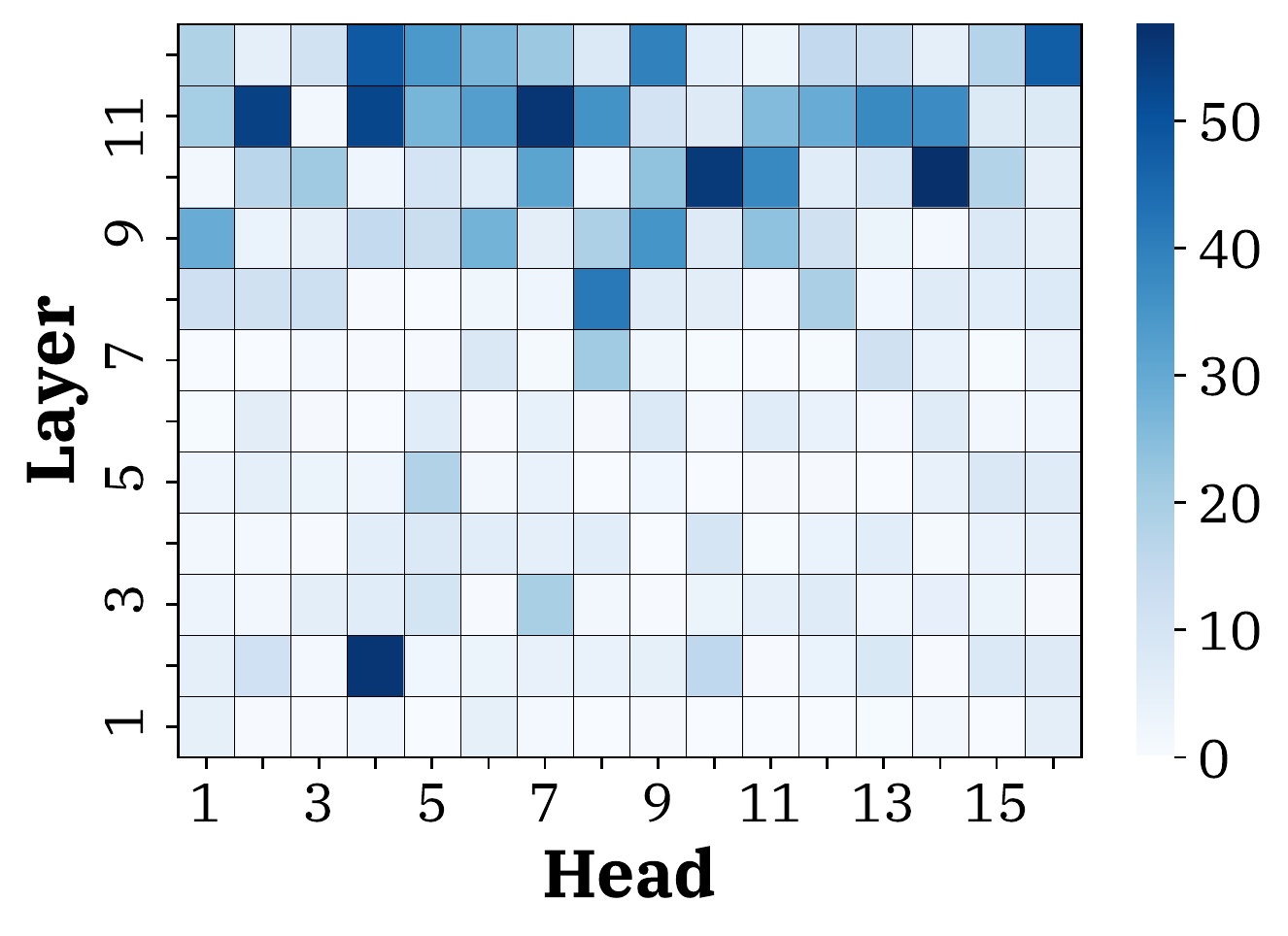}
         \caption{\textsc{copy content}}
     \end{subfigure}\hspace*{-0.4em}
     \begin{subfigure}[b]{0.24\textwidth}
         \centering
         \includegraphics[width=\textwidth]{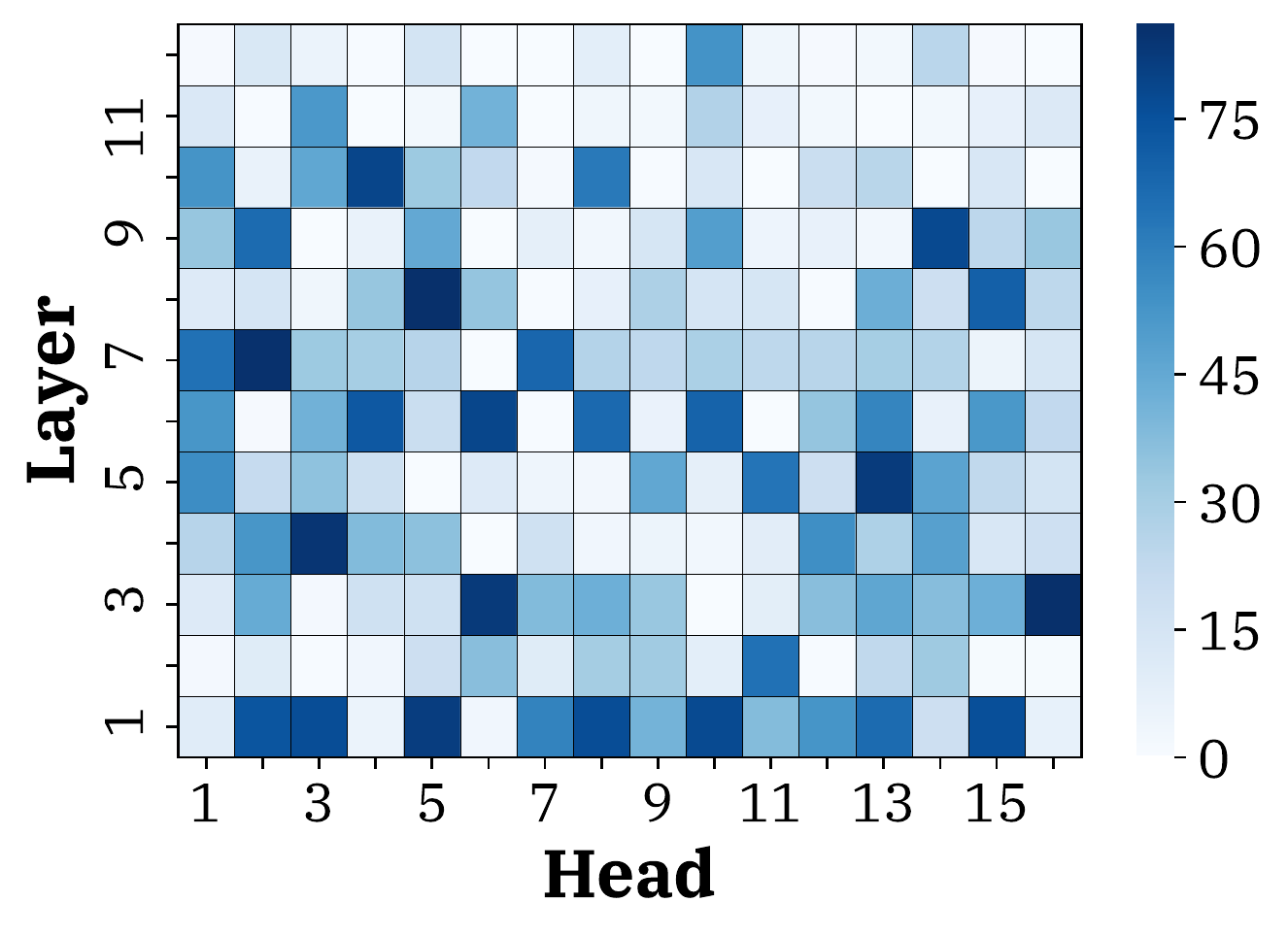}
         \caption{\textsc{first}}
     \end{subfigure}
    \caption{
     \textsc{copy} and \textsc{non-copy} \textsc{salient} attendee word percentages on the analysis set of CNN/DM. Top layers focus on words to be ``copied", while bottom layers attend to the broader salient context.}
    \label{fig:salient}
\end{figure}

We show the percentages of \textsc{copy} and \textsc{non-copy} \textsc{salient} attendees, \textsc{copy content} attendees, and \textsc{first} attendees on CNN/DM in Fig.~\ref{fig:salient}. As can be seen, top layers tend to focus on input tokens that will be generated as is, while bottom layers attend to salient words that are not used for current generation. Additionally, bottom layers frequently attend to the first token of the document, where bottom layers are more likely performing context gathering.
On NYT and XSum (figures are in Appendix~\ref{append:attention_focus}), similar trends are observed except that the \textsc{first} attendees are more focused by the top layers on NYT articles, where many of them start with all capitalized words.

\section{Summarization Results with Attention Head Masking}
\label{sec:exp_result}

In this section, we show how to leverage attention head masking and a content selector to improve summary informativeness on three datasets. We first train a binary sequence tagger for each dataset to label salient tokens in the source, used for \textbf{system masking} for attention heads. Our sequence tagger is a RoBERTa~\cite{liu2019roberta} encoder followed by a double layer multilayer perceptron (MLP) with a hyperbolic tangent activation function in between. To obtain the probability for each token, the MLP output is further fed into a sigmoid activation function. Details for training and decoding are in Appendix~\ref{append:training_and_decoding}.

The decision boundary for the sequence tagger is selected according to the F1 score calculated between the predicted tags and the ground-truth labels on the validation set. We search for the best decision boundary from $0.1$ to $0.4$, with a step size of $0.01$. The final decision boundaries used for taggers trained on CNN/DM, NYT, XSum are $0.20$, $0.24$, and $0.18$, achieving ROUGE-1 F1 of $43.70$, $44.10$, and $31.56$, respectively.

To select which heads at which layers to mask, we employ a greedy selection strategy. On the analysis set, we gradually apply system masking on four heads with most ROUGE improvement according to the study in \S~\ref{sec:content_selection}, and we select the heads that achieve the highest sum of ROUGE-1 F1 and ROUGE-2 F1. We apply four heads each time to reduce computational cost of hyperparameter searching. Heads selected for each dataset are in Appendix~\ref{append:head_selection}.

\begin{table}[t]
    \begin{subtable}[h]{0.48\textwidth}
    \small
    \setlength{\tabcolsep}{1.5pt}
    \hspace{-1mm}
        \begin{tabular}{p{0.6\textwidth}lll}
            \toprule
            \textbf{Model} & \textbf{R-1} & \textbf{R-2} & \textbf{R-L} \\
            \midrule
            \textsc{BERTSum}~\cite{liu-lapata-2019-text} & 42.13 & 19.60 & 39.18 \\
            \textsc{UniLM}~\cite{NIPS2019_9464} & 43.33 & 20.21 & 40.51 \\
            \textsc{PEGASUS}~\cite{zhang2019pegasus} & 44.17 & 21.47 & 41.11 \\
            \textsc{ProphetNet}~\citep{yan2020prophetnet} & 44.20 & 21.17 & 41.39 \\
            \hdashline
            \rule{-2pt}{2ex}
            BART (ours) & 44.19 & 21.20 & 40.98 \\
            \ + attention head masking (ours) & \textbf{45.54}{$^\ast$} & \textbf{22.24}{$^\ast$} & \textbf{42.44}{$^\ast$} \\
            \bottomrule
        \end{tabular}
        \vspace{-1mm}
        \caption{CNN/DailyMail}
        
    \end{subtable}
    \begin{subtable}[h]{0.48\textwidth}
    \small
    \setlength{\tabcolsep}{2.5pt}
    \hspace{-1mm}
        \begin{tabular}{p{0.6\textwidth}lll}
        \toprule
            \textbf{Model} & \textbf{R-1} & \textbf{R-2} & \textbf{R-L} \\
            \midrule
            \textsc{BottomUp}~\cite{gehrmann-etal-2018-bottom} & 47.38 & 31.23 & 41.81 \\
            \textsc{DCA}~\cite{celikyilmaz-etal-2018-deep} & 48.08 & 31.19 & 42.33 \\
            \textsc{BERTSum}~\cite{liu-lapata-2019-text} & 49.02 & 31.02 & 45.55 \\
            \textsc{ASGARD}~\cite{huang2020knowledge} & 51.29 & 34.97 & 48.26 \\
            \hdashline\rule{-2pt}{2ex}
            BART (ours) & 53.00 & 36.31 & 48.90 \\
            \ + attention head masking (ours) & \textbf{53.52}{$^\ast$} & \textbf{36.69} & \textbf{49.24} \\
            \bottomrule
        \end{tabular}
        \vspace{-1mm}
        \caption{New York Times}
    \end{subtable}
    
    \begin{subtable}[h]{0.48\textwidth}
    \small
    \setlength{\tabcolsep}{3pt}
    \centering
        \begin{tabular}{p{0.6\textwidth}lll}
        \toprule
            \textbf{Model} & \textbf{R-1} & \textbf{R-2} & \textbf{R-L} \\
            \midrule
            \textsc{BERTSum}~\cite{liu-lapata-2019-text} & 38.81 & 16.50 & 31.27 \\
            \textsc{PEGASUS} & \textbf{47.21} & \textbf{24.56} & \textbf{39.25} \\
            \hdashline\rule{-2pt}{2ex}
            BART (ours) & 45.36 & 22.30 & 37.11 \\
            \ + attention head masking (ours) & 45.35 & 22.31 & 37.15 \\
            \bottomrule
        \end{tabular}
        \vspace{-1mm}
        \caption{XSum}

    \end{subtable}
    
    \caption{
    Automatic evaluation with ROUGE.
    $\ast$: significantly better than BART with approximate randomization test ($p < 0.005$). 
    Our method outperforms BART and previous models on CNN/DM and NYT.
    }
    \label{tab:doc_summ}
\end{table}

\smallskip
\noindent \textbf{In-domain Results.} 
Table~\ref{tab:doc_summ} shows that applying our attention head masking technique on BART obtains significantly better results on CNN/DM and NYT, compared to several top performing abstractive summarization models trained with large Transformers.
The improvement is more pronounced for CNN/DM than the other two datasets. We believe this is due to the difference in abstractiveness among the three datasets. CNN/DM has more extractive summaries compared to the other datasets~\cite{grusky-etal-2018-newsroom}, suggesting attention head masking is more effective on extractive datasets.
Notably, \textsc{PEGASUS} is pre-trained with 3.8TB of news articles, the BART model used in our work is only pre-trained with 160GB of a combination of news, books, stories, and web text. The large size of the pre-training data might be a big contributor to the better performance by PEGASUS on XSum.

For \textbf{human evaluation}, we hire three fluent English speakers to rate $50$ pairs of summaries generated with and without attention head masking based on BART for \textbf{informativeness} and \textbf{faithfulness}. Informativeness measures how well the summary captures salient content from the article, while faithfulness indicates whether the summary correctly reflects the content in the source article. 
The annotators are asked to determine if attention head masking improves any of the two aspects. 
As shown in Table~\ref{tab:human_eval} where all ratings by three judges are considered, summaries generated with attention head masking are considered to have better informativeness, but no substantial improvement on faithfulness is observed. 

\begin{table}[t]
    \centering
    \small
    \setlength{\tabcolsep}{4pt}
    \begin{tabular}{lccc}
    \toprule
         & \textbf{w/ masking} & \textbf{w/o masking} & \textbf{Tie} \\
        \cmidrule(lr){2-3}
        \cmidrule(lr){4-4}
        \textbf{Informativeness} & \textbf{36.0\%} & 19.3\% & 44.7\% \\
        \textbf{Faithfulness} & \textbf{10.0\%} & 7.3\% & 82.7\% \\
        \bottomrule
    \end{tabular}
    \caption{Percentages of summaries with and without attention head masking favored by annotators on informativeness and faithfulness. The Krippendorff's $\alpha$ for informativeness and faithfulness are $0.30$ and $0.47$.
    }
    \label{tab:human_eval}
\end{table}

\begin{figure}[t]
    \centering
    \includegraphics[width=0.48\textwidth]{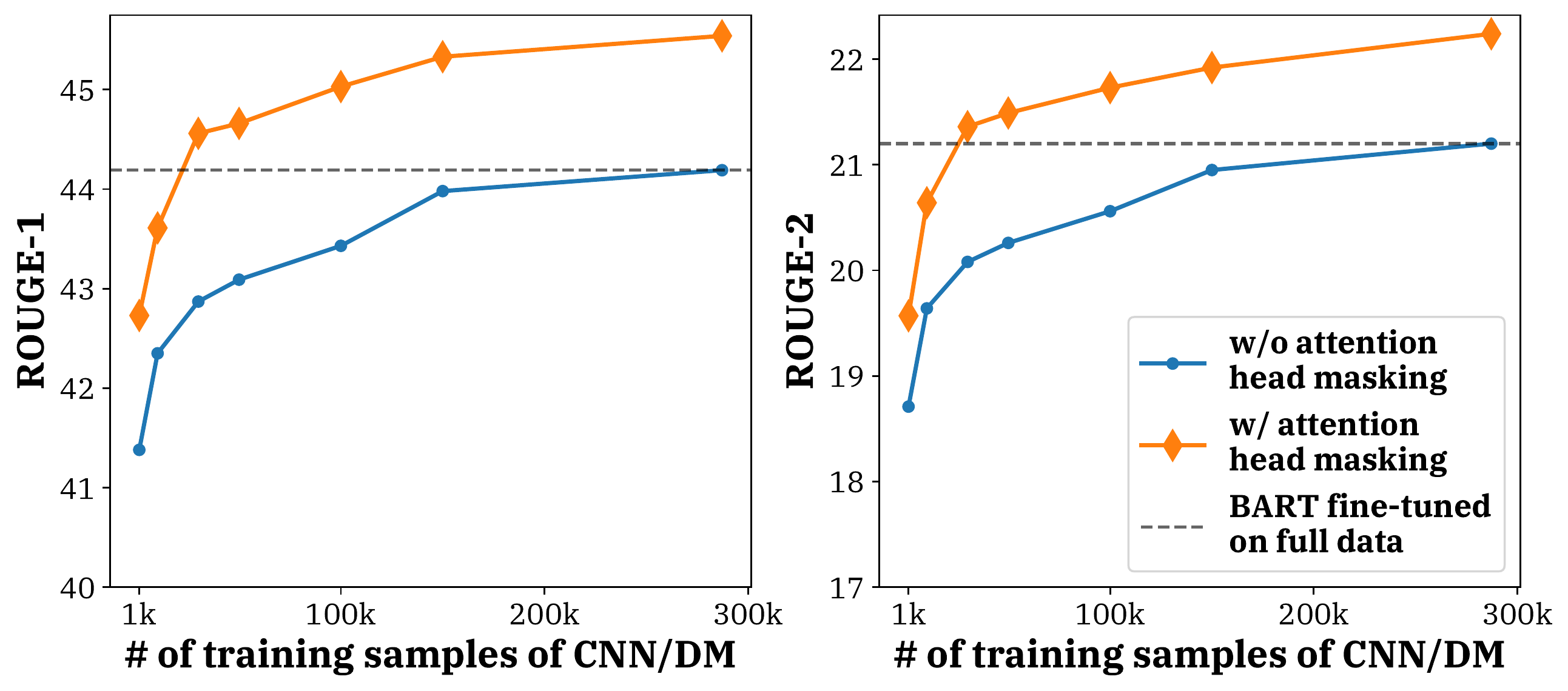}
    \caption{Results on CNN/DM with different sizes of training data. Our method consistently improves the summarizer.}
    \label{fig:different_cnndm_size}
\end{figure}

\smallskip
\noindent \textbf{Limited Training Data.} 
Next, we study if our masking technique is still effective if given limited training samples. We use the limited training samples to train both the summarizer and the content selector.
As can be seen in Fig.~\ref{fig:different_cnndm_size}, our masking technique consistently increases ROUGE scores with varying amounts of training data. Notably, our model trained on only $30$K samples (with attention head masking) outperforms the model trained on the full dataset, suggesting that directly informing content selection is more data-efficient than model fine-tuning on more summaries.

\smallskip
\noindent \textbf{Cross-domain Results.} 
Finally, we show results on NYT using BART fine-tuned on CNN/DM, with system masks predicted by a tagger trained on different sizes of NYT samples (Table~\ref{tab:cross_domain}). Using a selector trained with only $10$k of target domain samples, we already significantly improve the performance by BART trained on CNN/DM only.

\begin{table}[t]
    \centering
    \small
    \begin{tabular}{lccc}
    \toprule
        \textbf{Selector Training Data} & \textbf{R-1} & \textbf{R-2} & \textbf{R-L} \\
        \midrule
        No masking & 31.11 & 14.68 & 28.19 \\
        $10$K & 34.98 & 17.95 & 31.87 \\
        $100$K & 34.71 & 17.70 & 31.61 \\
        $589$K (full) & 35.13 & 18.07 & 32.03 \\
        \bottomrule
    \end{tabular}
    \caption{Results on NYT summaries generated by BART trained on CNN/DM, with masks predicted by content selectors trained on different sizes of NYT data.}
    \label{tab:cross_domain}
\end{table}

\section{Conclusion}

We propose attention head masking that constrains encoder-decoder attention heads to attend to salient tokens, to inform content selection in abstractive summarization. 
With this technique, we first demonstrate the relation between encoder-decoder attentions and content selection behaviors. 
With system masks predicted by external content selectors, we show that attention head masking can consistently improve ROUGE scores over competitive summarization models on three benchmarks.
Summaries generated with attention head masking are also preferred by human judges more frequently.
Additional experiments demonstrate that our method is more data-efficient and effective on both in-domain and cross-domain settings.

\section*{Acknowledgements}
This research is supported in part by National Science Foundation through Grant IIS-1813341, and by the Office of the Director of National Intelligence (ODNI), Intelligence Advanced Research Projects Activity (IARPA), via contract \# FA8650-17-C-9116. The views and conclusions contained herein are those of the authors and should not be interpreted as necessarily representing the official policies, either expressed or implied, of ODNI, IARPA, or the U.S. Government. The U.S. Government is authorized to reproduce and distribute reprints for governmental purposes notwithstanding any copyright annotation therein. 
We thank three anonymous reviewers for their constructive suggestions. 

\newpage

\bibliography{custom}
\bibliographystyle{acl_natbib}

\appendix

\section{Training and Decoding Settings}\label{append:training_and_decoding}

When training the sequence taggers, we minimize the average binary cross-entropy of each token's selection probability relative to the ground-truth label. The parameters of the RoBERTa encoder are fixed.
We set the learning rate to $5 \times 10^{-4}$ and batch size to $128$. Unless specified, all the models in this paper are trained with Adam~\cite{DBLP:journals/corr/KingmaB14} optimizer and training will be stopped if there is no improvement on the validation set for $2$ consecutive epochs.

For BART models, we follow the instructions provided by Fairseq~\cite{ott2019fairseq} to set the training hyperparameters on CNN/DM and XSum. We use the same hyperparameters for CNN/DM and NYT, except that we adopt a linear learning rate decay of $30{,}000$ steps in total for NYT.

During test, we use a beam size of $5$, $5$, $6$ for CNN/DM, NYT, and XSum, respectively. To reduce computational cost, we use beam size $1$ for our analysis experiments on all datasets. The length penalties are $2.0$, $1.5$ and $1.0$ for CNN/DM, NYT, and XSum, following \citet{lewis-etal-2020-bart}. We set the minimal and maximal lengths during decoding as: $55$ and $140$ for CNN/DM, $0$ and $140$ for NYT, and $10$ and $60$ for XSum.

\section{Head Selection}\label{append:head_selection}

For CNN/DM, we apply masking to all heads at layer 1. The ROUGE-1/2/L F1 on the analysis set are $36.43/16.02/33.59$.

For NYT, we apply masking to 12 heads at layer 3. The indices of heads are: 3, 4, 6, 7, 8, 9, 10, 11, 12, 13, 14, 15. The ROUGE-1/2/L F1 on the analysis set are $55.27/39.20/48.16$.

For XSum, we apply masking to 12 heads at layer 3. The indices of heads are: 1, 2, 3, 4, 6, 7, 8, 9, 11, 13, 14, 15. The ROUGE-1/2/L F1 on the analysis set are $45.77/22.82/37.60$.

\section{Content Selection Effects on XSum and NYT}\label{append:content_selection_effect}

\begin{figure}[t]
    \centering
    \begin{subfigure}[b]{0.45\textwidth}
         \centering
         \includegraphics[width=\textwidth]{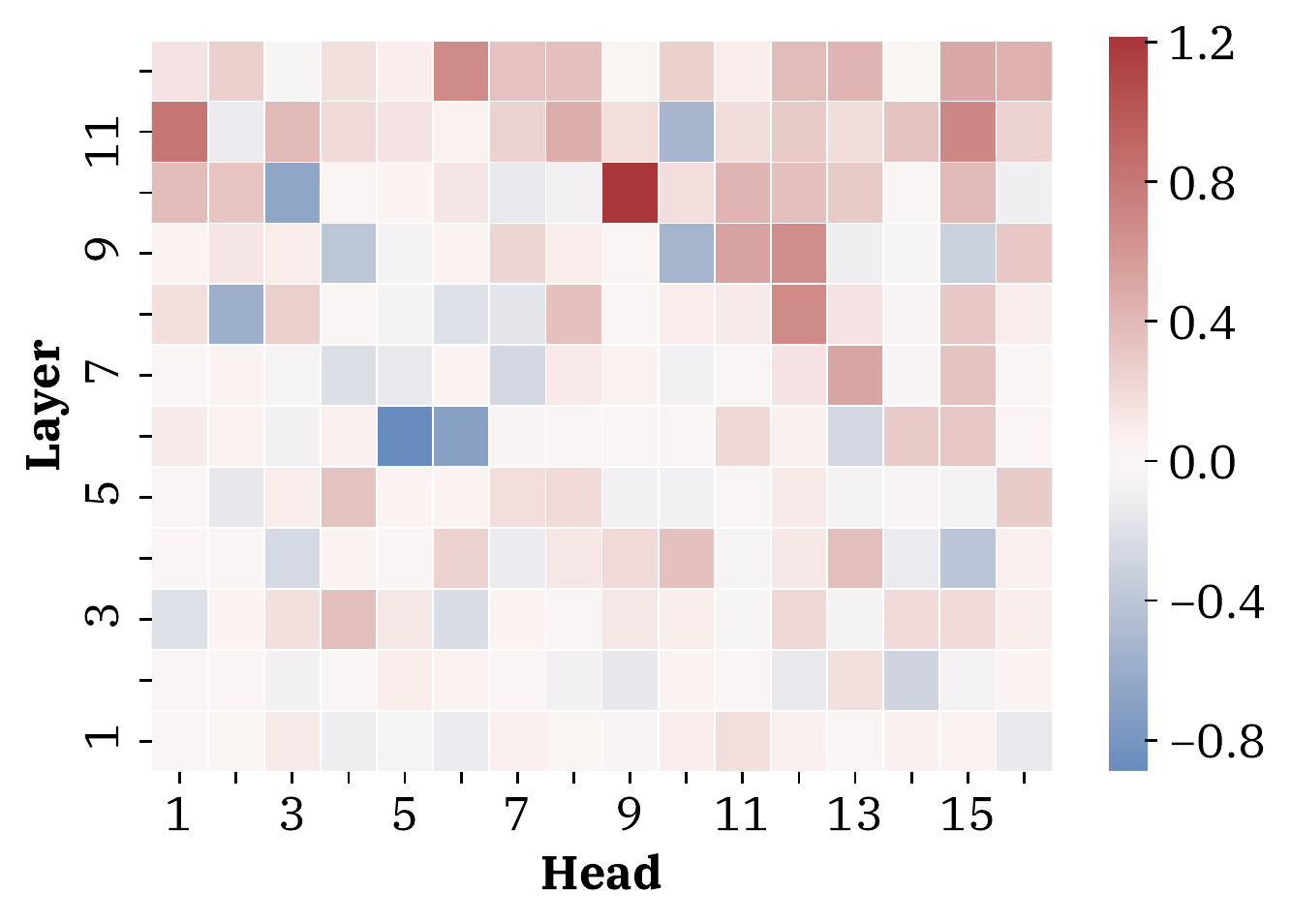}
         \caption{XSum}
     \end{subfigure}
     \begin{subfigure}[b]{0.45\textwidth}
         \centering
         \includegraphics[width=\textwidth]{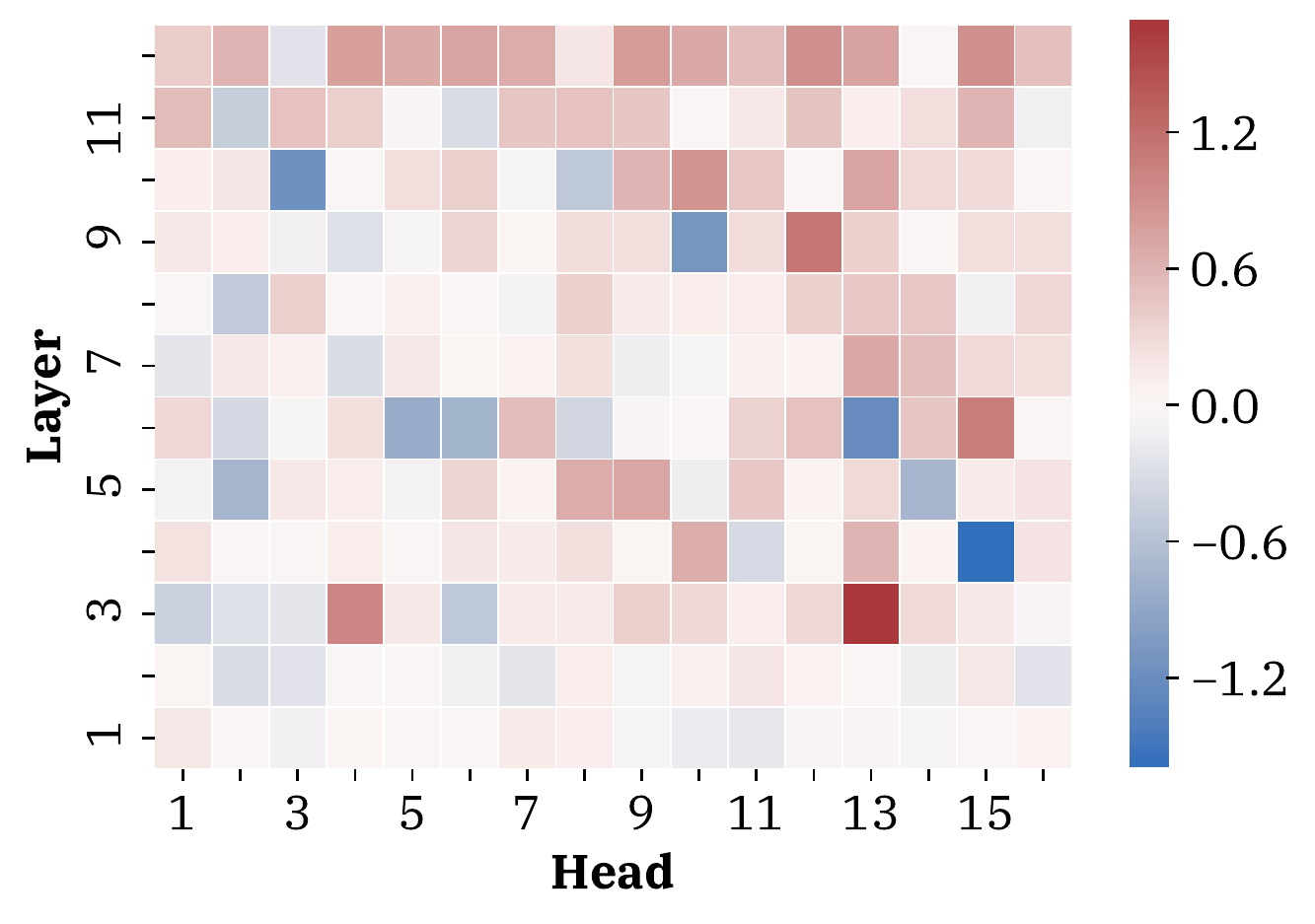}
         \caption{NYT}
     \end{subfigure}
    \caption{ROUGE-1 improvement with oracle masks for each head at each layer on the analysis sets of XSum and NYT.}
    \label{fig:head_perf}
\end{figure}

The content selection effects for BART models fine-tuned on XSum and NYT, measured by the ROUGE improvement from the uniform attention weight setting to the oracle masking setting, are shown in Fig.~\ref{fig:head_perf}.

On all three datasets, it is more effective to constrain attentions to salient content at the top layers. Especially, the top layer yields the most ROUGE-1 improvement. Moreover, the ROUGE improvement by a specific head varies among different datasets.

\section{Additional Results for Synergism Analysis}\label{append:synergism_analysis}

\begin{figure}[t]
    \centering
    \begin{subfigure}[b]{0.48\textwidth}
         \centering
         \includegraphics[width=\textwidth]{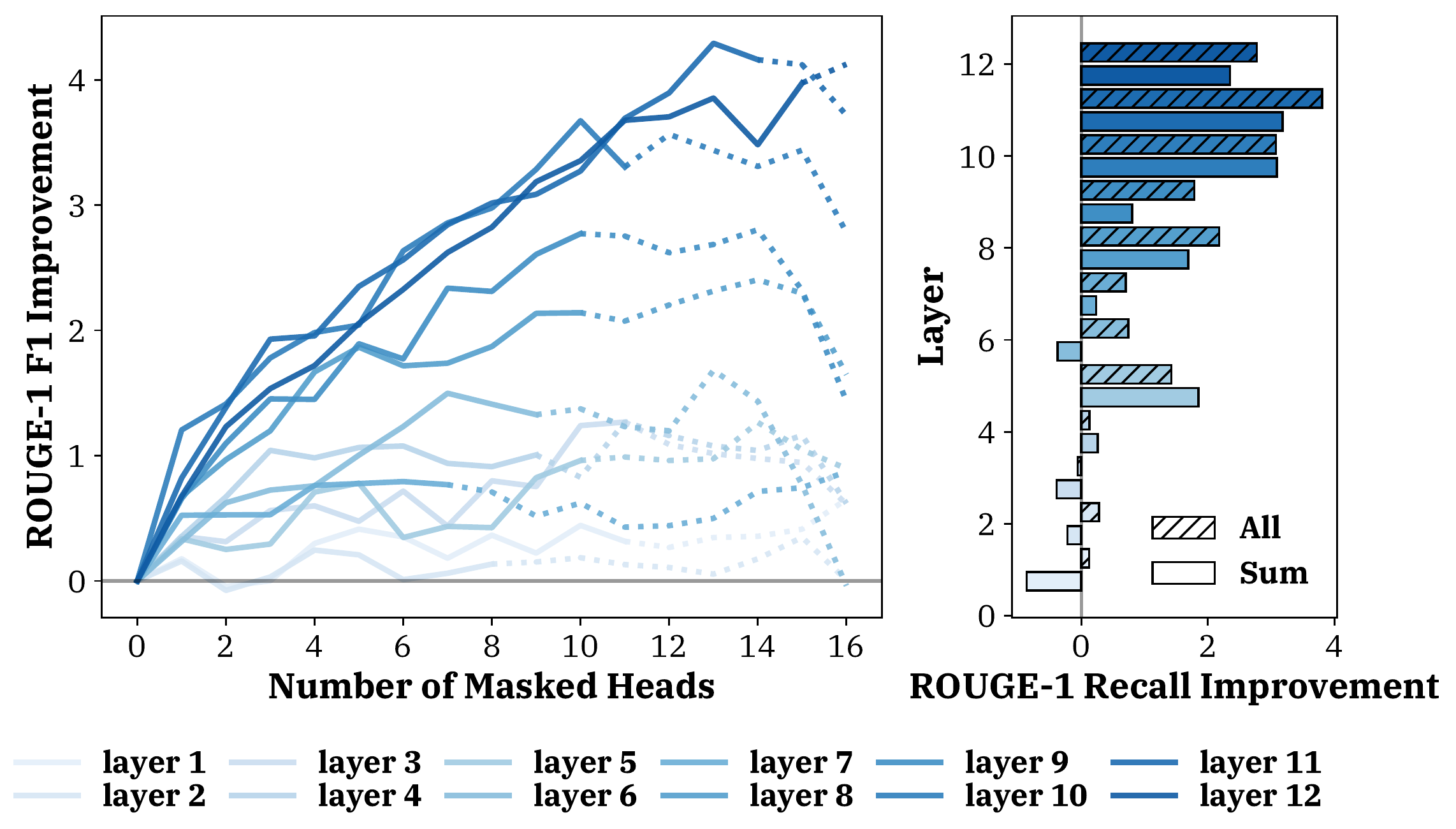}
         \caption{XSum}
     \end{subfigure}
     \begin{subfigure}[b]{0.48\textwidth}
         \centering
         \includegraphics[width=\textwidth]{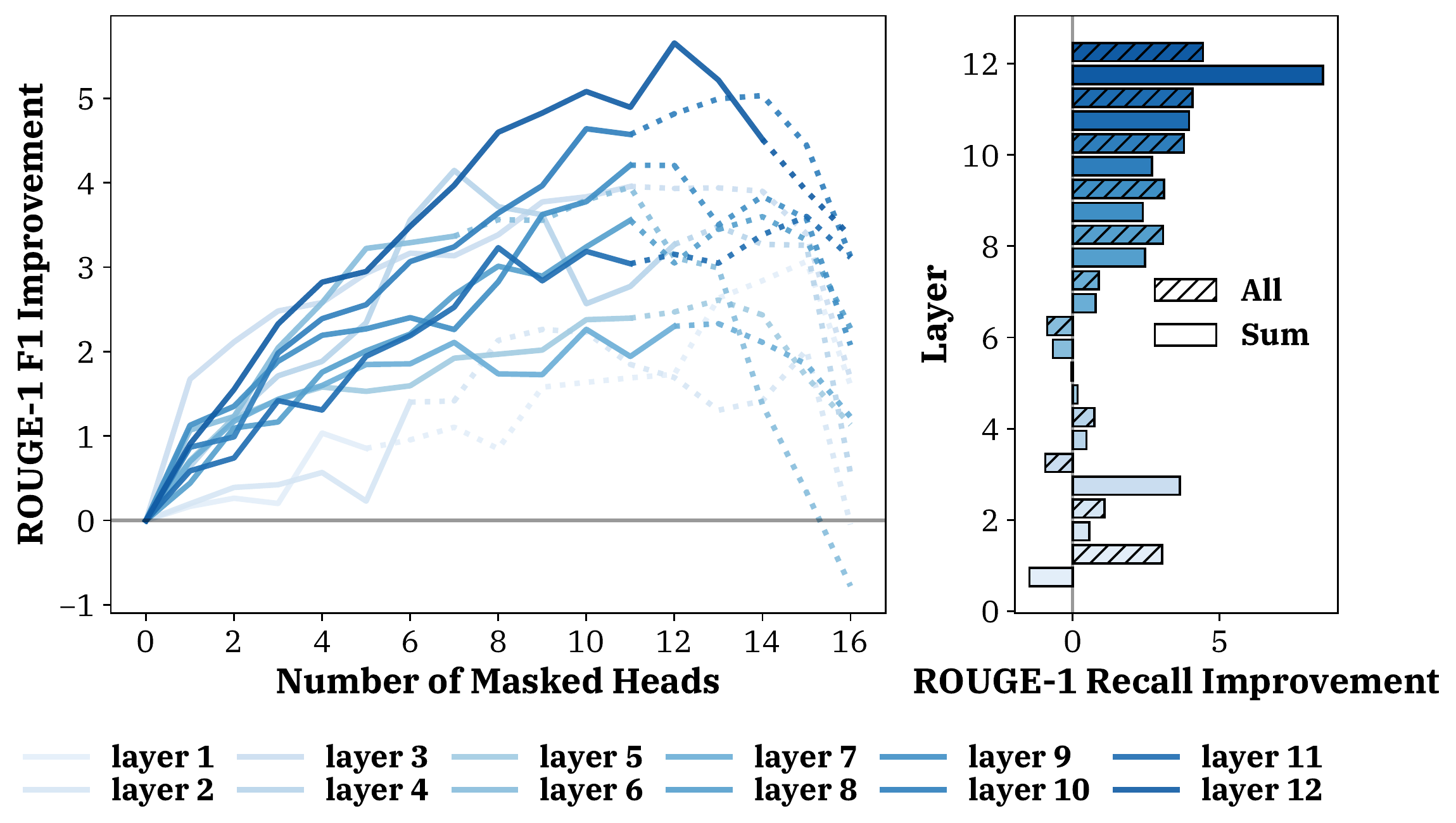}
         \caption{NYT}
     \end{subfigure}
    \caption{[\textbf{Left}] ROUGE-1 F1 improvement by incrementally applying oracle masking to the next head with most ROUGE improvement per layer on XSum and NYT. Dotted lines indicate that the newly masked heads do not have individual ROUGE improvements. 
    [\textbf{Right}] ROUGE-1 recall improvement by masking all heads vs. sum of improvement by masking each head separately on XSum and NYT.
    Better displayed with color.
    }
    \vspace{-3mm}
    \label{fig:synergy}
\end{figure}

We show the synergism analysis for models fine-tuned on XSum and NYT in Fig.~\ref{fig:synergy}. They both echo the observation on CNN/DM that multiple heads have strong collaborative effects and heads may not be independent at pinpointing different salient content.

\section{Attention Focus}\label{append:attention_focus}

\begin{figure}[ht]
     \centering
     \begin{subfigure}[b]{0.24\textwidth}
         \centering
         \includegraphics[width=\textwidth]{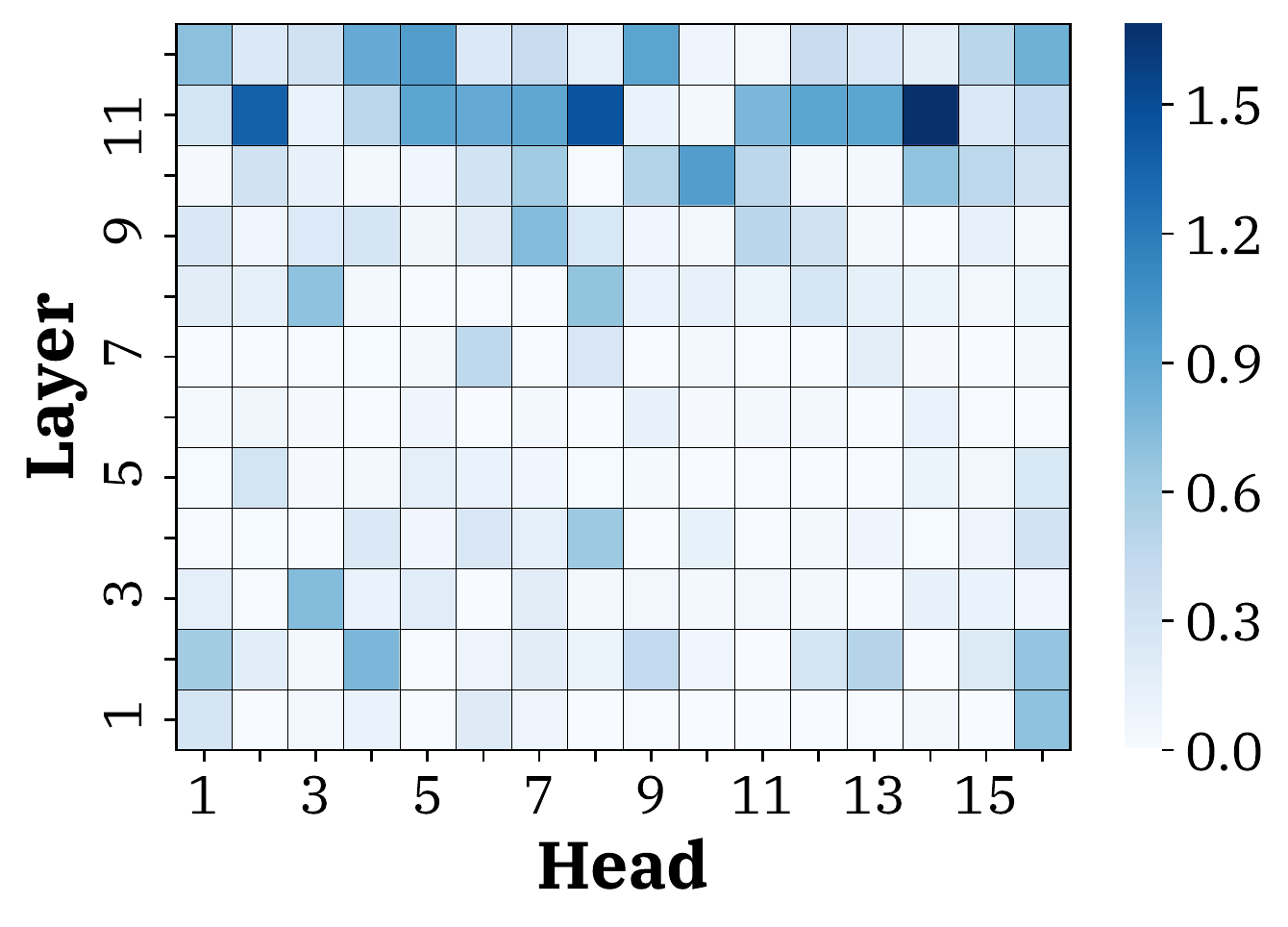}
         \caption{\textsc{copy salient}}
     \end{subfigure}\hspace*{-0.4em}
     \begin{subfigure}[b]{0.24\textwidth}
         \centering
         \includegraphics[width=\textwidth]{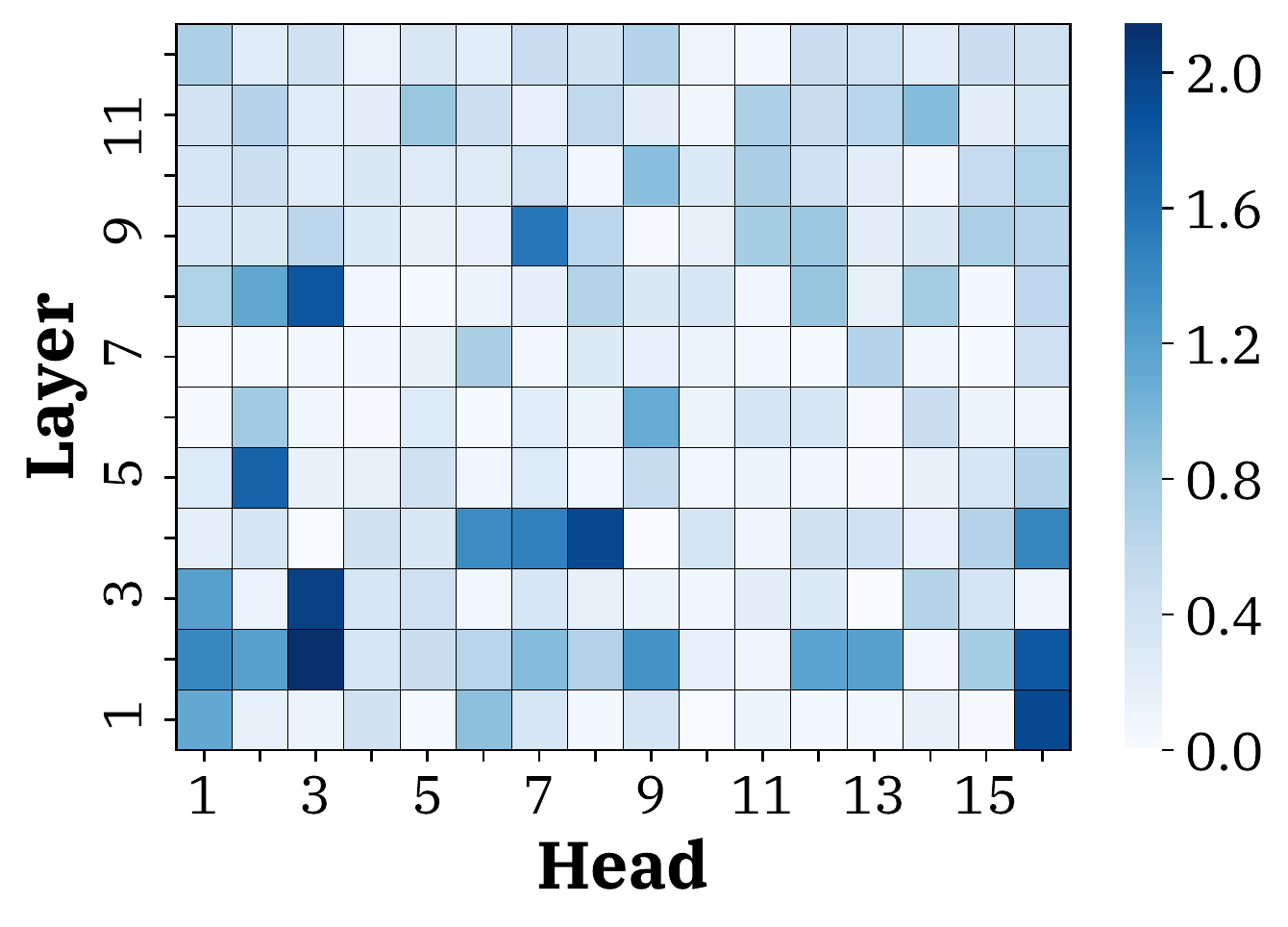}
         \caption{\textsc{non-copy salient}}
     \end{subfigure}
     \begin{subfigure}[b]{0.24\textwidth}
         \centering
         \includegraphics[width=\textwidth]{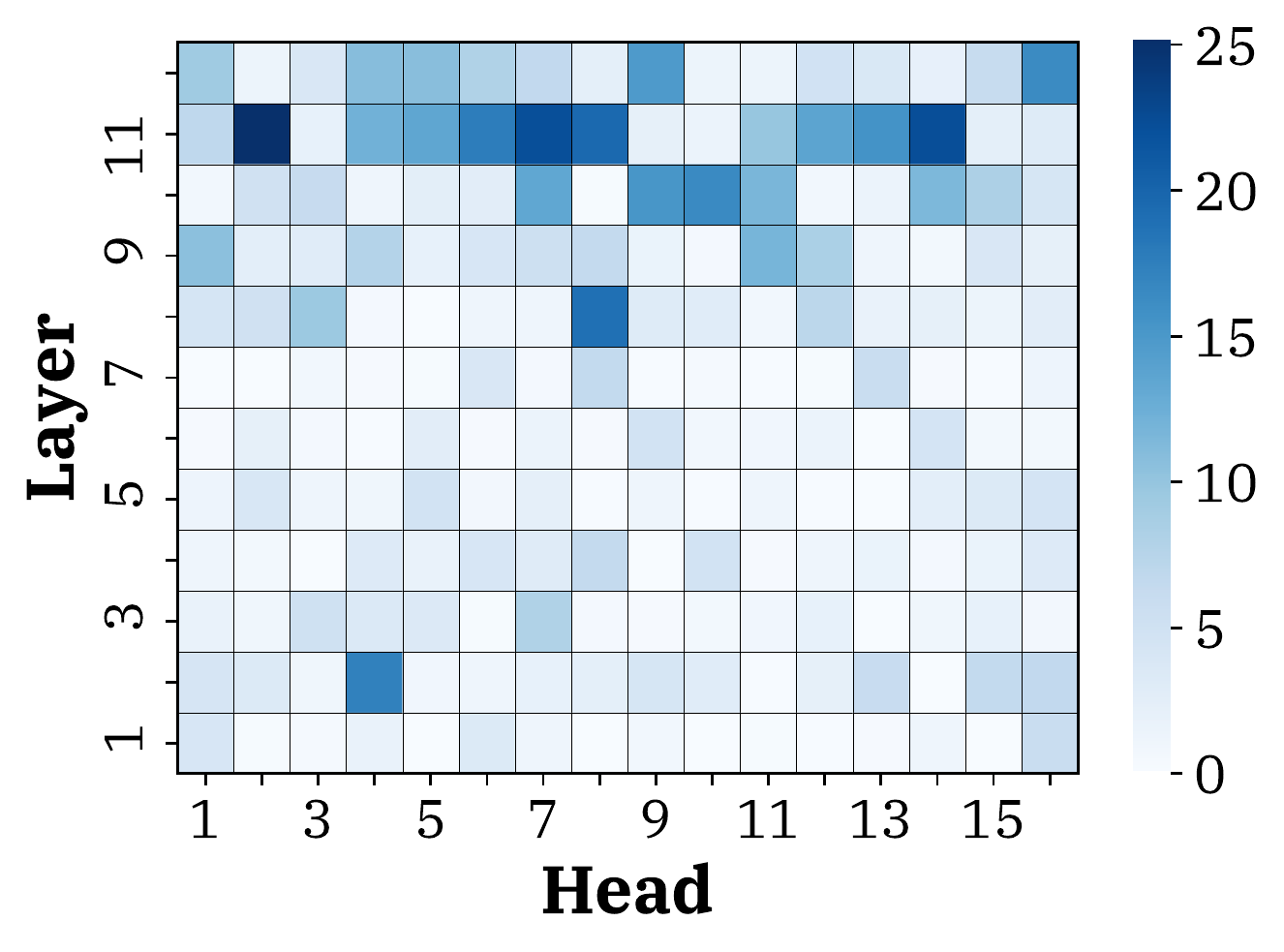}
         \caption{\textsc{copy content}}
     \end{subfigure}\hspace*{-0.4em}
     \begin{subfigure}[b]{0.24\textwidth}
         \centering
         \includegraphics[width=\textwidth]{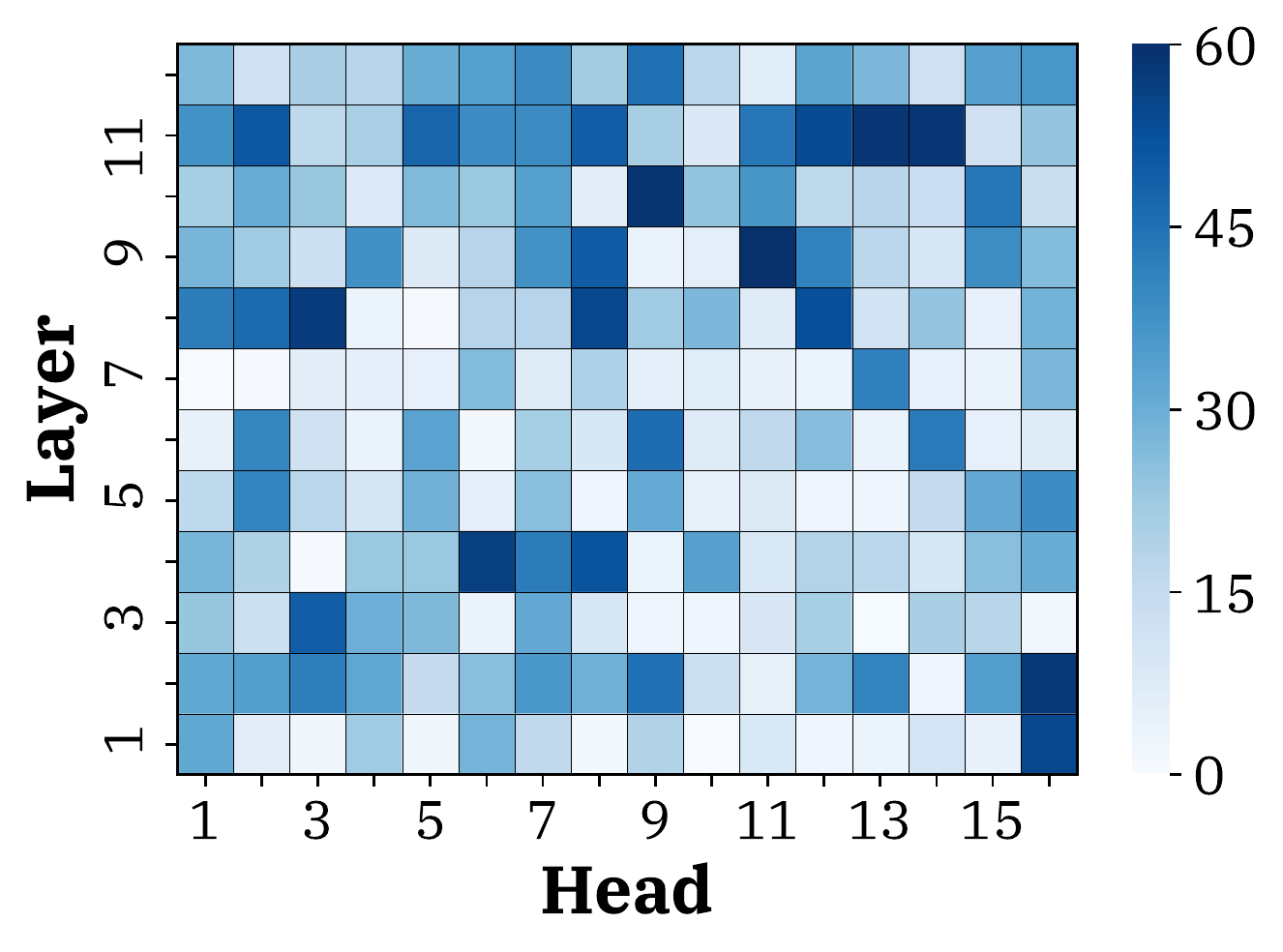}
         \caption{\textsc{non-copy content}}
     \end{subfigure}
     \begin{subfigure}[b]{0.24\textwidth}
         \centering
         \includegraphics[width=\textwidth]{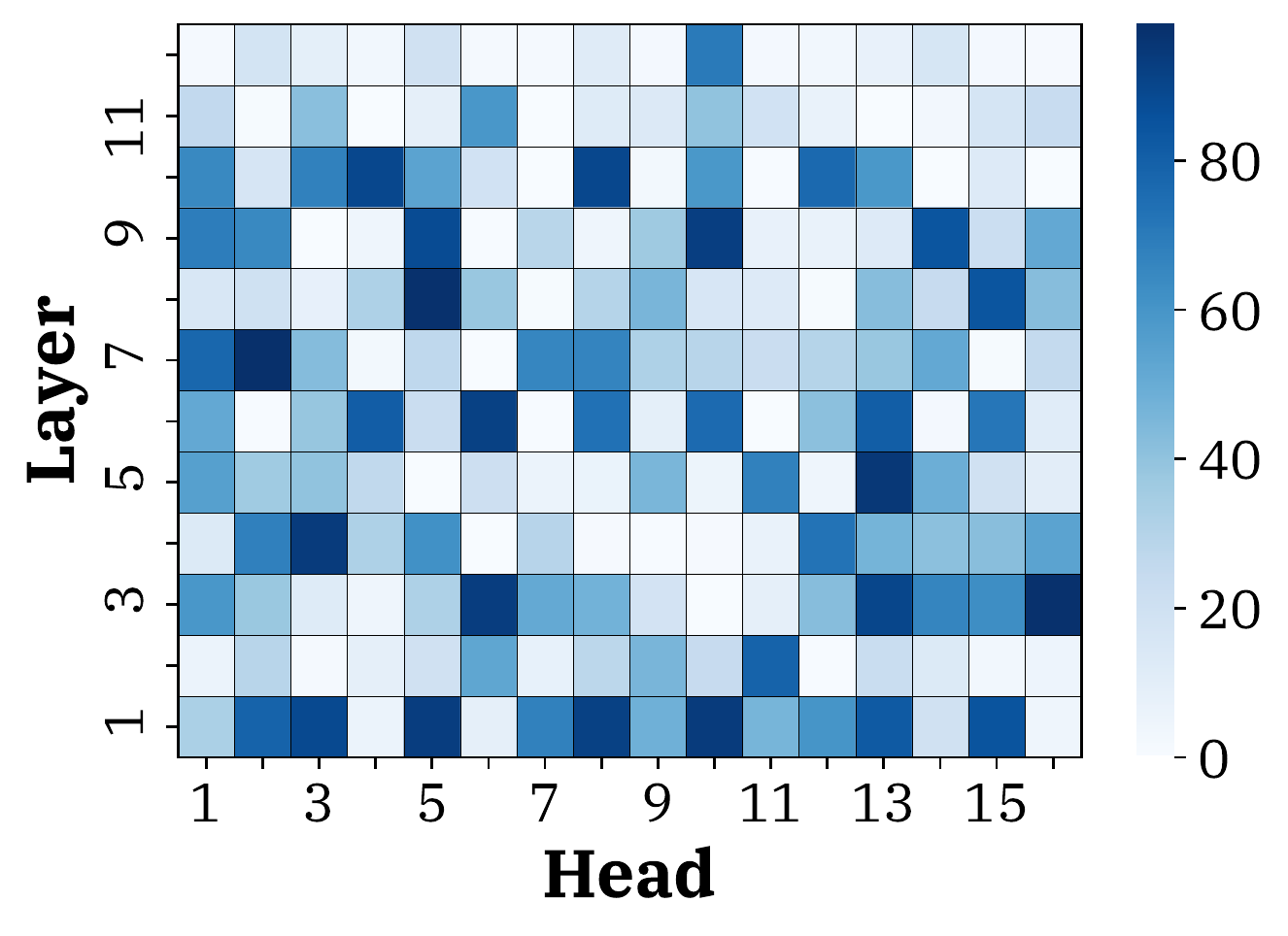}
         \caption{\textsc{first}}
     \end{subfigure}\hspace*{-0.4em}
     \begin{subfigure}[b]{0.24\textwidth}
         \centering
         \includegraphics[width=\textwidth]{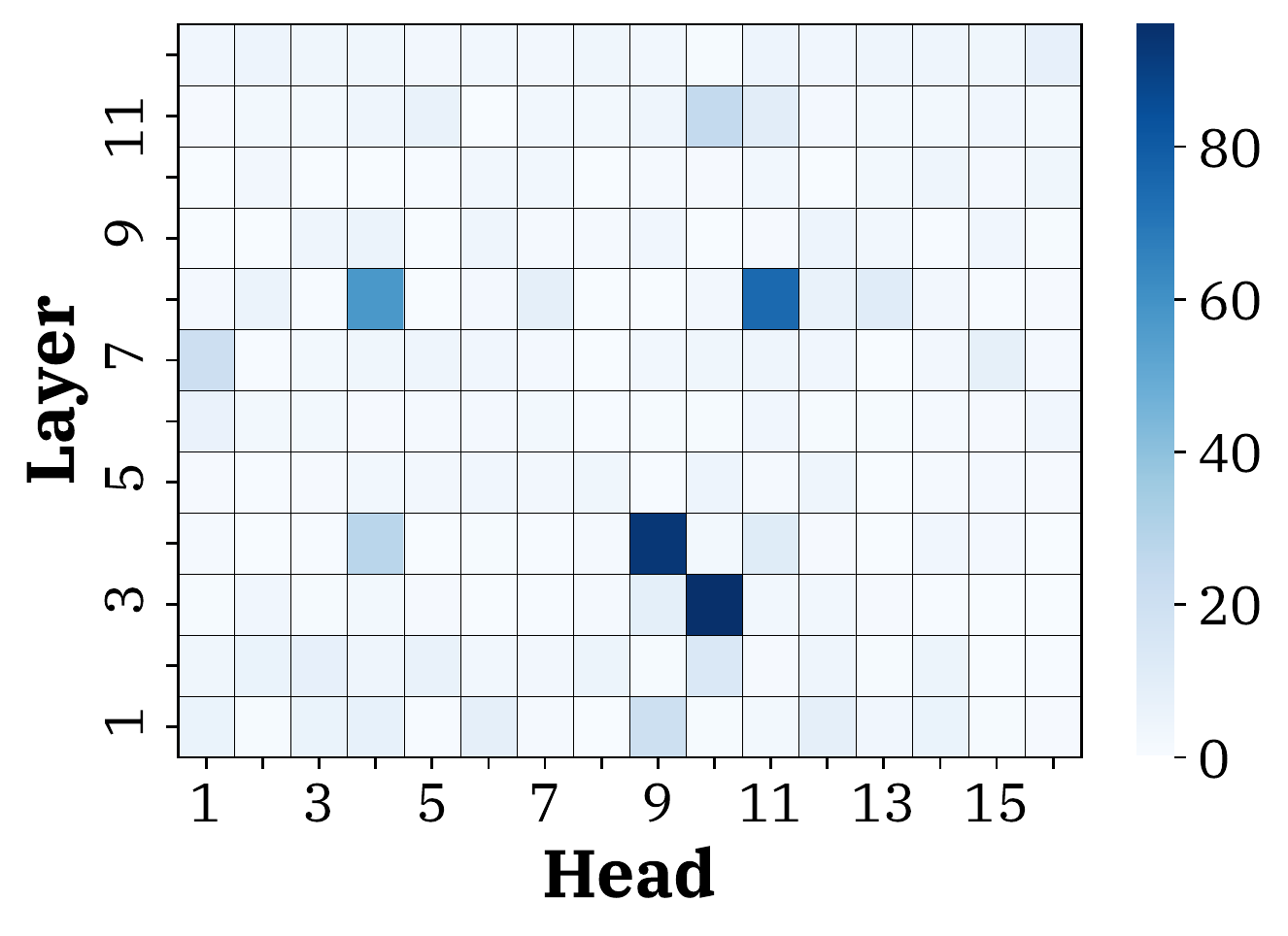}
         \caption{\textsc{last}}
     \end{subfigure}
    \caption{
    Percentages of \textsc{copy salient}, \textsc{non-copy salient}, \textsc{copy content}, \textsc{non-copy content}, \textsc{first} and \textsc{last} attendees for each head at each layer on the analysis set of XSum.}
    \label{fig:xsum_focus}
\end{figure}

\begin{figure}[!t]
     \centering
     \begin{subfigure}[b]{0.24\textwidth}
         \centering
         \includegraphics[width=\textwidth]{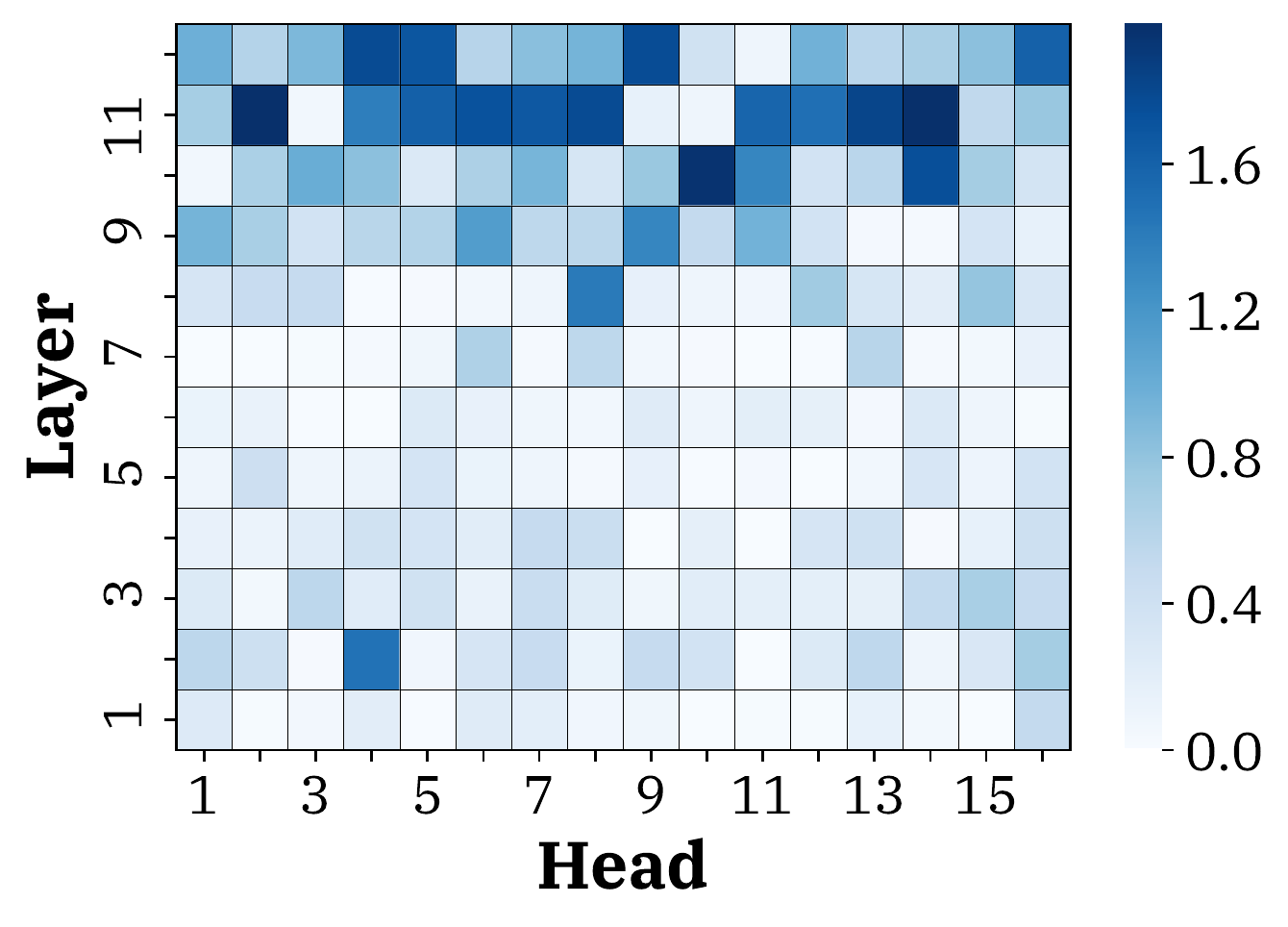}
         \caption{\textsc{copy salient}}
     \end{subfigure}\hspace*{-0.4em}
     \begin{subfigure}[b]{0.24\textwidth}
         \centering
         \includegraphics[width=\textwidth]{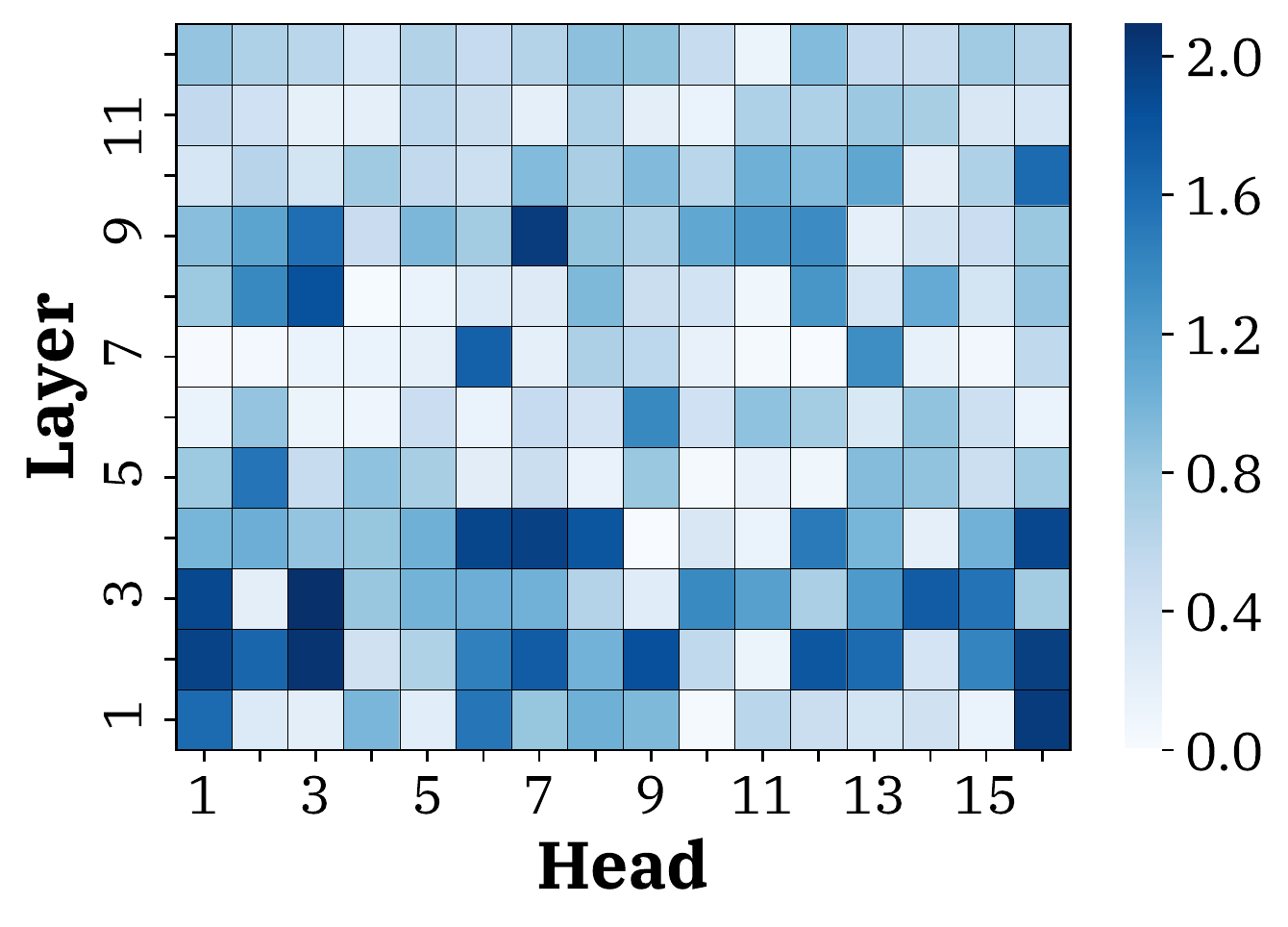}
         \caption{\textsc{non-copy salient}}
     \end{subfigure}
     \begin{subfigure}[b]{0.24\textwidth}
         \centering
         \includegraphics[width=\textwidth]{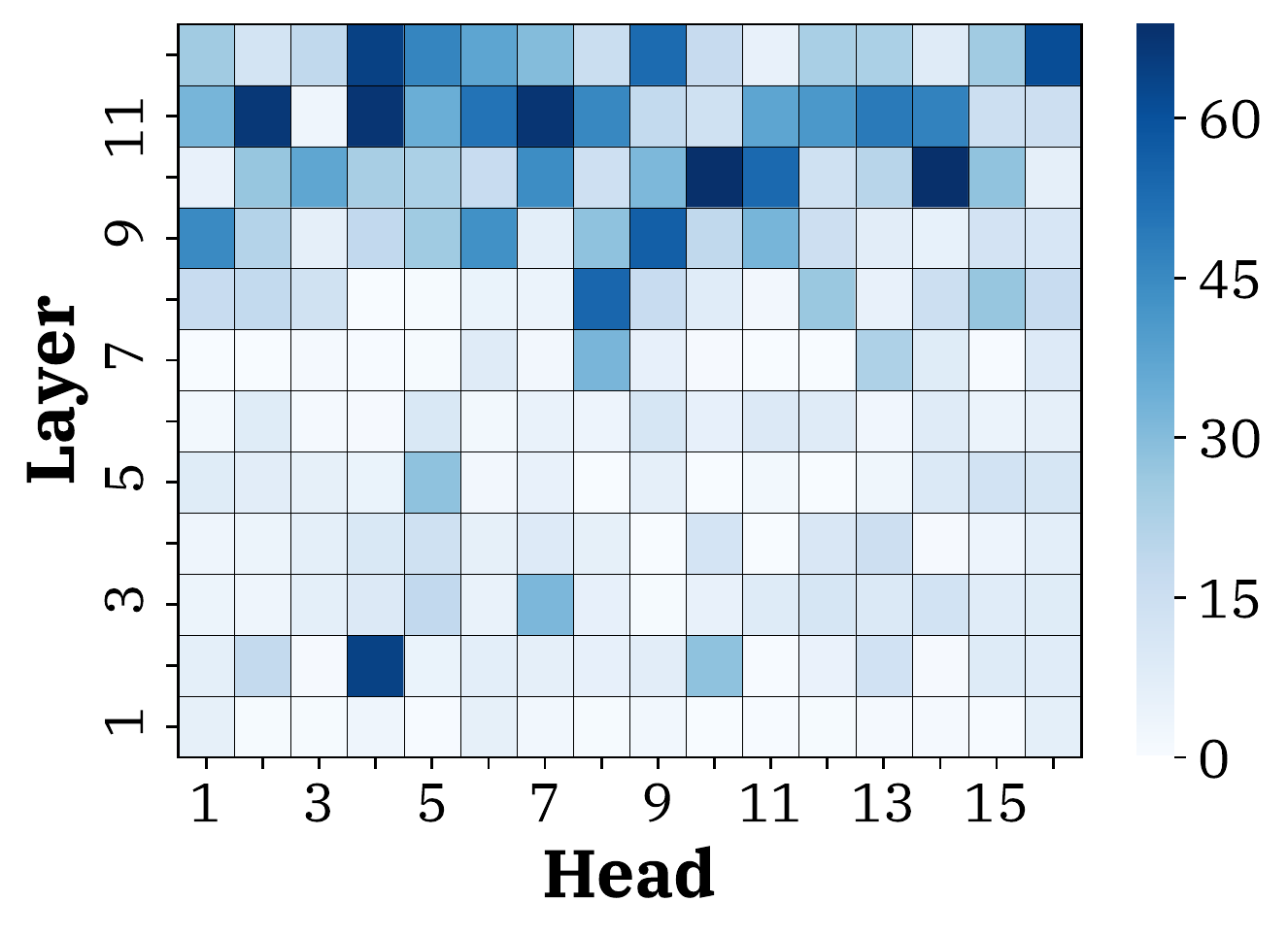}
         \caption{\textsc{copy content}}
     \end{subfigure}\hspace*{-0.4em}
     \begin{subfigure}[b]{0.24\textwidth}
         \centering
         \includegraphics[width=\textwidth]{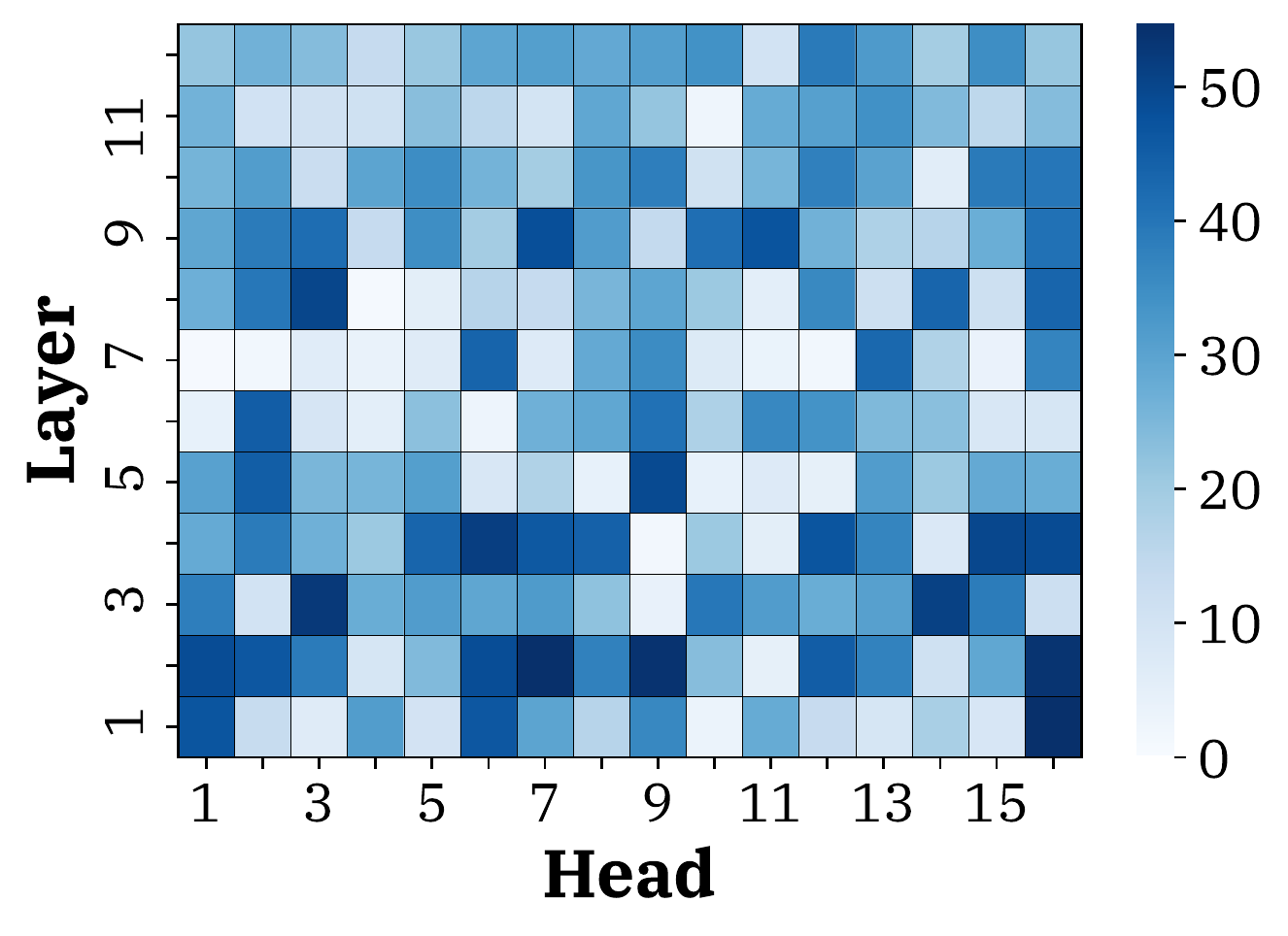}
         \caption{\textsc{non-copy content}}
     \end{subfigure}
     \begin{subfigure}[b]{0.24\textwidth}
         \centering
         \includegraphics[width=\textwidth]{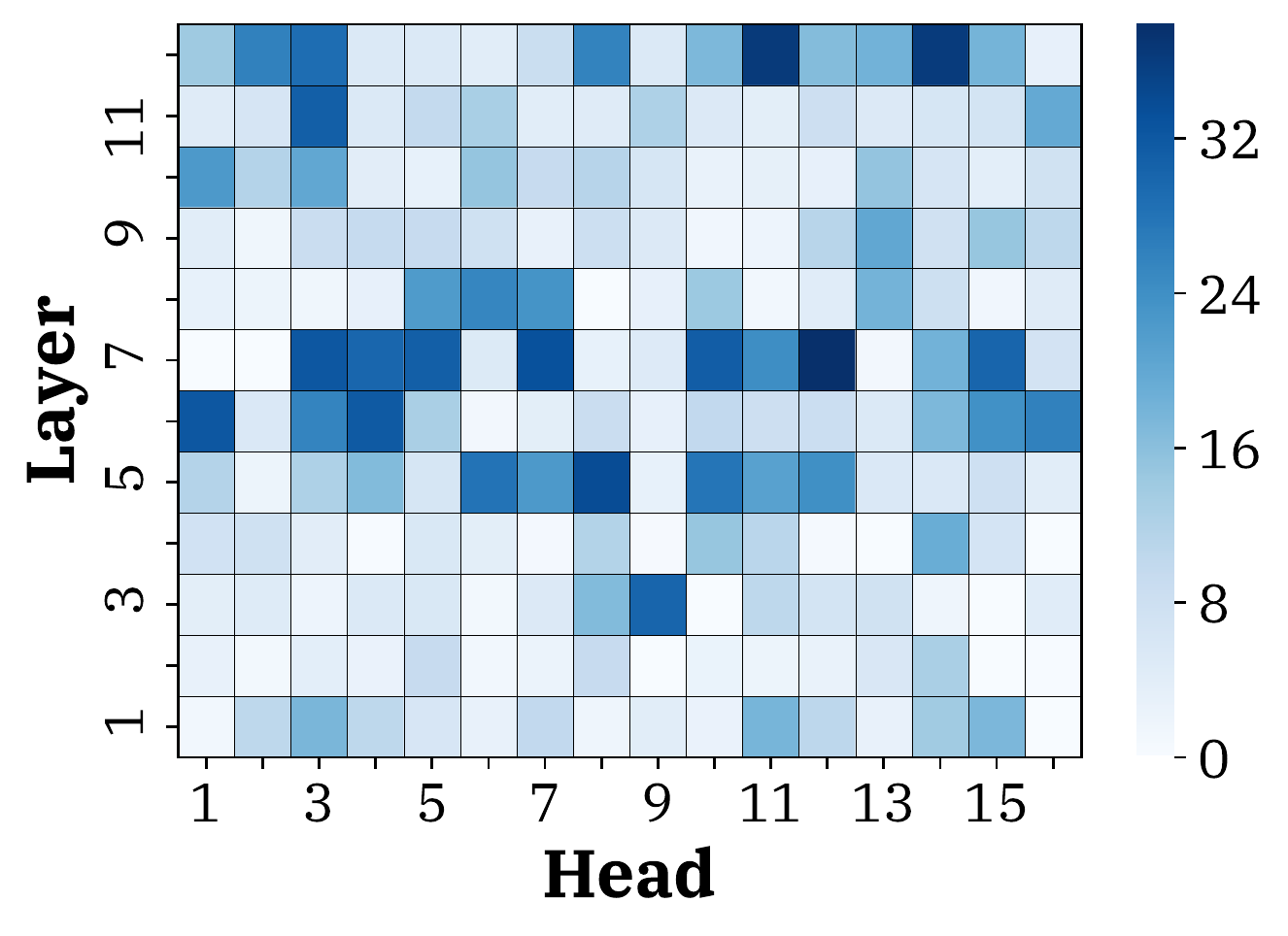}
         \caption{\textsc{first}}
     \end{subfigure}\hspace*{-0.4em}
     \begin{subfigure}[b]{0.24\textwidth}
         \centering
         \includegraphics[width=\textwidth]{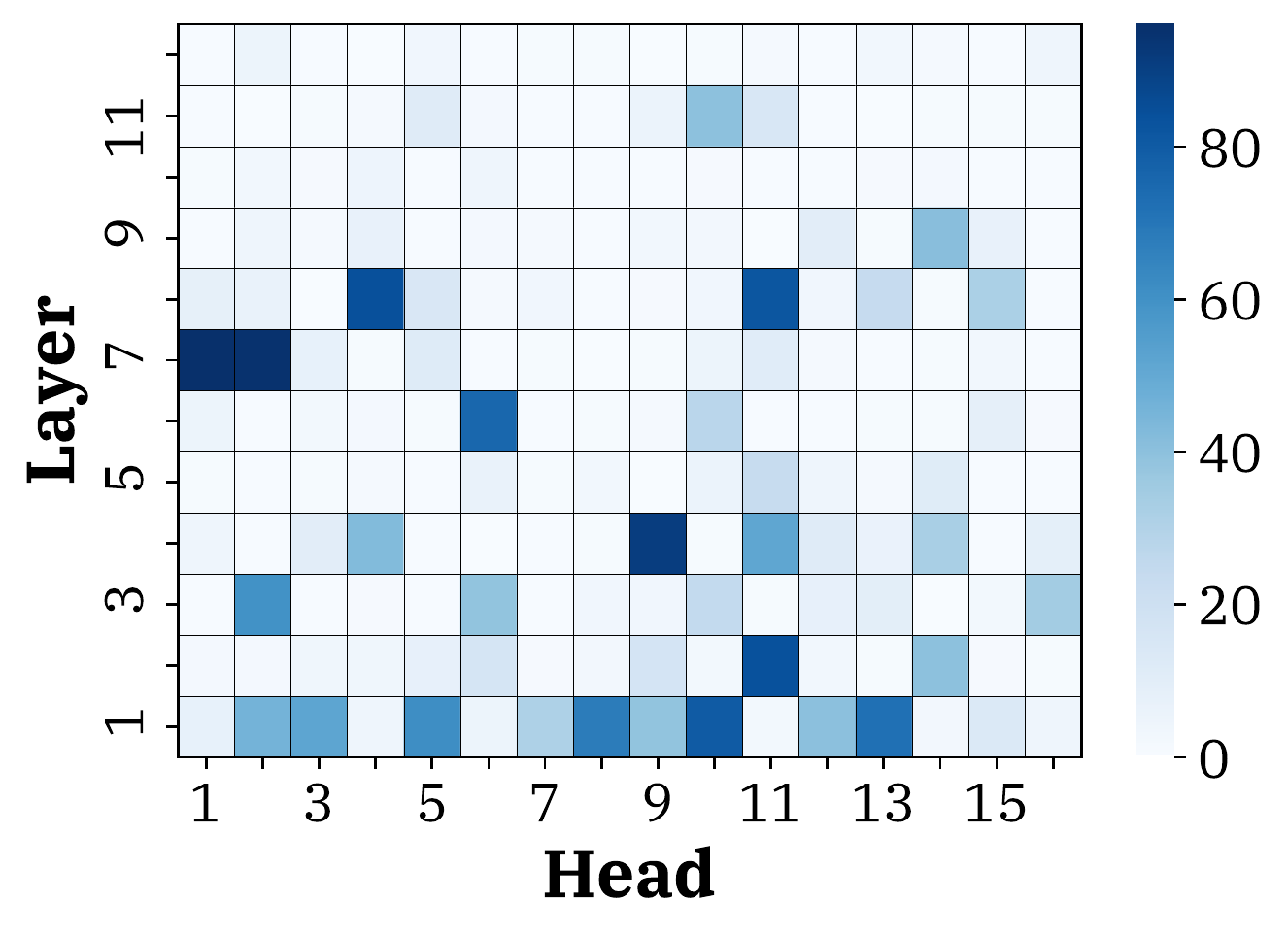}
         \caption{\textsc{last}}
     \end{subfigure}
    \caption{
    Percentages of \textsc{copy salient}, \textsc{non-copy salient}, \textsc{copy content}, \textsc{non-copy content}, \textsc{first} and \textsc{last} attendees for each head at each layer on the analysis set of NYT.}
    \label{fig:nyt_focus}
\end{figure}

\begin{figure}[t]
     \centering
     \begin{subfigure}[b]{0.24\textwidth}
         \centering
         \includegraphics[width=\textwidth]{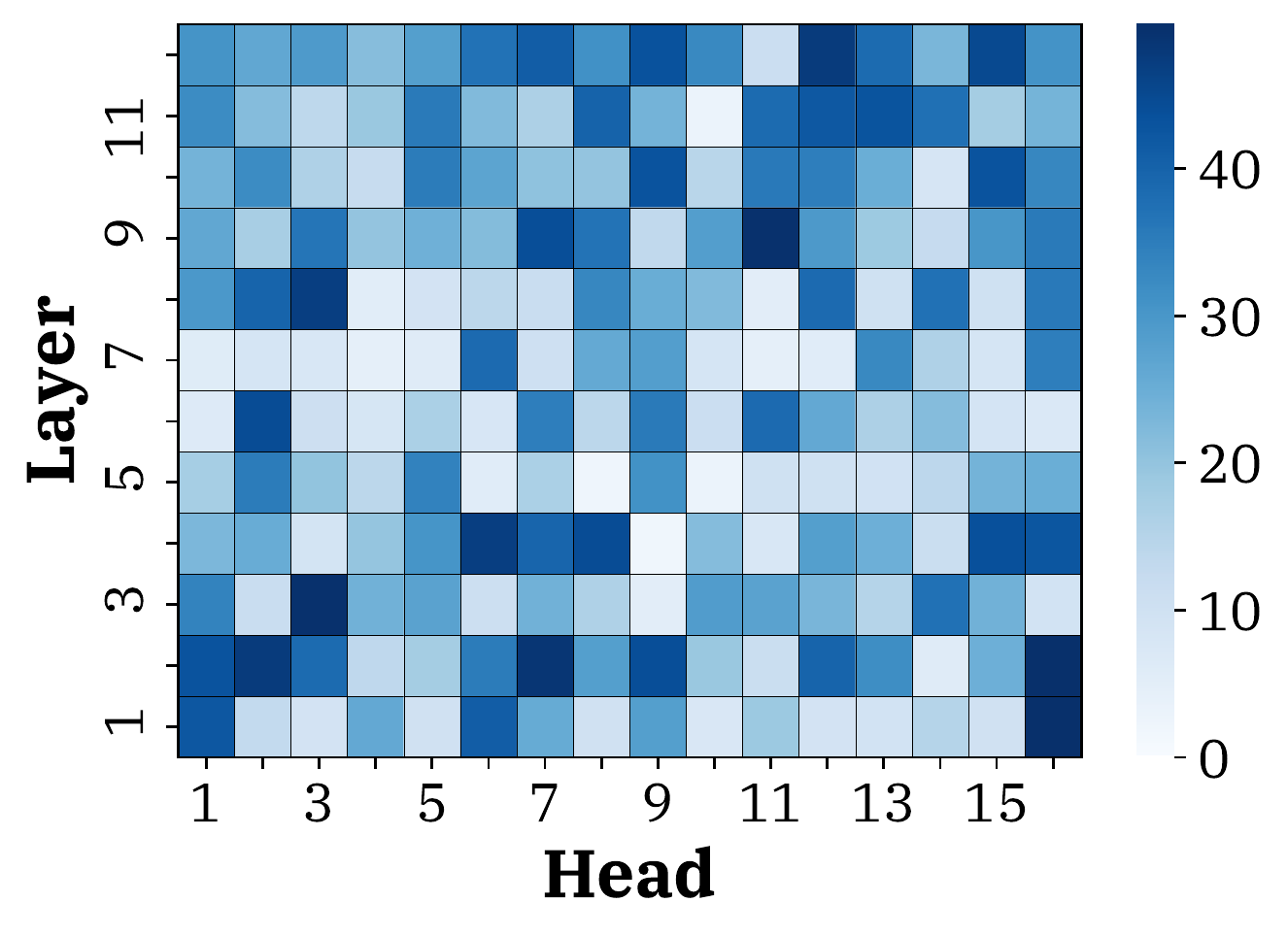}
         \caption{\textsc{non-copy content}}
     \end{subfigure}\hspace*{-0.4em}
     \begin{subfigure}[b]{0.24\textwidth}
         \centering
         \includegraphics[width=\textwidth]{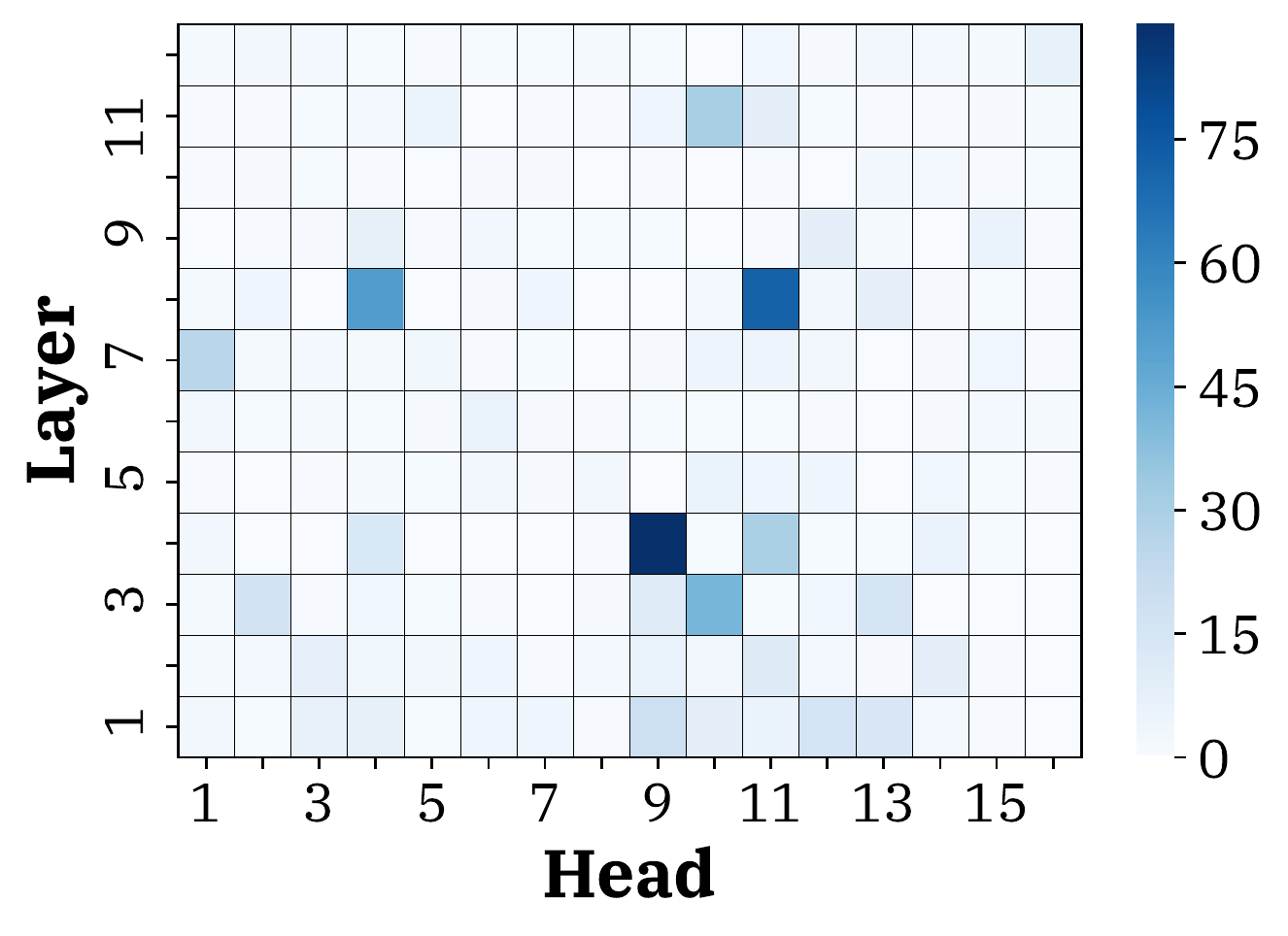}
         \caption{\textsc{last}}
     \end{subfigure}
    \caption{
    Percentages of \textsc{non-copy content} and \textsc{last} attendees for each head at each layer on the analysis set of CNN/DM.
    }
    \label{fig:cnndm_focus}
    \vspace{10in}
\end{figure}

We show the percentages of each type of attendees on the analysis sets of XSum, NYT, and CNN/DM in Fig.~\ref{fig:xsum_focus}, Fig.~\ref{fig:nyt_focus}, and Fig.~\ref{fig:cnndm_focus}, respectively. We find that heads have similar focus for salient words, content words, and the last word across different datasets. 
Interestingly, the attention focus for the first word on NYT is different from other datasets. On NYT, many articles start with all capitalized words, which might become the focus of some heads.

\end{document}